\documentclass[conference]{IEEEtran}
\IEEEoverridecommandlockouts
\usepackage{cite}
\usepackage{amsmath,amssymb,amsfonts}
\usepackage{graphicx}
\usepackage{algorithm}
\usepackage{algpseudocode}
\usepackage{algorithmicx}
\usepackage[dvipsnames]{xcolor}
\usepackage{tabularx}
\usepackage{booktabs}
\usepackage[russian, english]{babel}
\usepackage[T2A]{fontenc}   
\usepackage[utf8]{inputenc} 
\usepackage{subcaption} 

\usepackage{float}
\usepackage[unicode]{hyperref}
\def\BibTeX{{\rm B\kern-.05em{\sc i\kern-.025em b}\kern-.08em
    T\kern-.1667em\lower.7ex\hbox{E}\kern-.125emX}}

\begin{document}

\bibliographystyle{IEEEtran}

\title{COLIBRI Fuzzy Model: Color Linguistic-Based Representation and Interpretation%
\thanks{Corresponding author: Pakizar Shamoi (e-mail: p.shamoi@kbtu.kz).}%
\thanks{This research was funded by the Science Committee of the Ministry of Science and Higher Education of the Republic of Kazakhstan (Grant No. AP22786412).}%
}

\author{
\IEEEauthorblockN{
Pakizar Shamoi\IEEEauthorrefmark{1}, Nuray Toganas\IEEEauthorrefmark{1}, Muragul Muratbekova\IEEEauthorrefmark{1},
Elnara Kadyrgali\IEEEauthorrefmark{1}, Adilet Yerkin\IEEEauthorrefmark{1}, Ayan Igali\IEEEauthorrefmark{1},\\
Malika Ziyada\IEEEauthorrefmark{1}, Ayana Adilova\IEEEauthorrefmark{1}, Aron Karatayev\IEEEauthorrefmark{1}, Yerdauit Torekhan\IEEEauthorrefmark{1}
}
\IEEEauthorblockA{\IEEEauthorrefmark{1}School of Information Technology and Engineering, Kazakh-British Technical University, Almaty, Kazakhstan}
}


%
%
%
%
%
%
%
%
%




\maketitle


\begin{abstract}
Colors are omnipresent in today's world and play a vital role in how humans perceive and interact with their surroundings. However, it is challenging for computers to imitate human color perception. This paper introduces the Human Perception-Based Fuzzy Color Model, COLIBRI (Color Linguistic-Based Representation and Interpretation), designed to bridge the gap between computational color representations and human visual perception. The proposed model uses fuzzy sets and logic to create a framework for color categorization. Using a three-phase experimental approach, the study first identifies distinguishable color stimuli for hue, saturation, and intensity through preliminary experiments, followed by a large-scale human categorization survey involving more than 1000 human subjects. The resulting data are used to extract fuzzy partitions and generate membership functions that reflect real-world perceptual uncertainty. The model incorporates a mechanism for adaptation that allows refinement based on feedback and contextual changes. Comparative evaluations demonstrate the model's alignment with human perception compared to traditional color models, such as RGB, HSV, and LAB. To the best of our knowledge, no previous research has documented the construction of a model for color attribute specification based on a sample of this size or a comparable sample of the human population (n = 2496). Our findings are significant for fields such as design, artificial intelligence, marketing, and human-computer interaction, where perceptually relevant color representation is critical.

\end{abstract}


\section{Introduction} 
Colors play a fundamental role in how we interpret, experience, and interact with the world, influencing everything from emotions to decision-making. It is integral to AI, design, architecture, marketing, and psychological functioning \cite{2019COLORISTICS, Jaglarz2023Perception, ProLab2021, yan2022rgb}. However, accurately replicating human color perception remains a major challenge for computational models \cite{burambekova2024comparative}. Existing color models, such as RGB, HSV, and LAB, are widely used in digital applications; however, they fail to capture the gradual and context-dependent nature of human perception. These models rely on rigid boundaries and fixed categories, which do not align with how humans perceive color transitions, particularly under varying lighting conditions, cultural contexts, or emotional states \cite{Schwarz1987An}.

\begin{figure}[tb]
  \centering

 \includegraphics[width=0.5\textwidth]{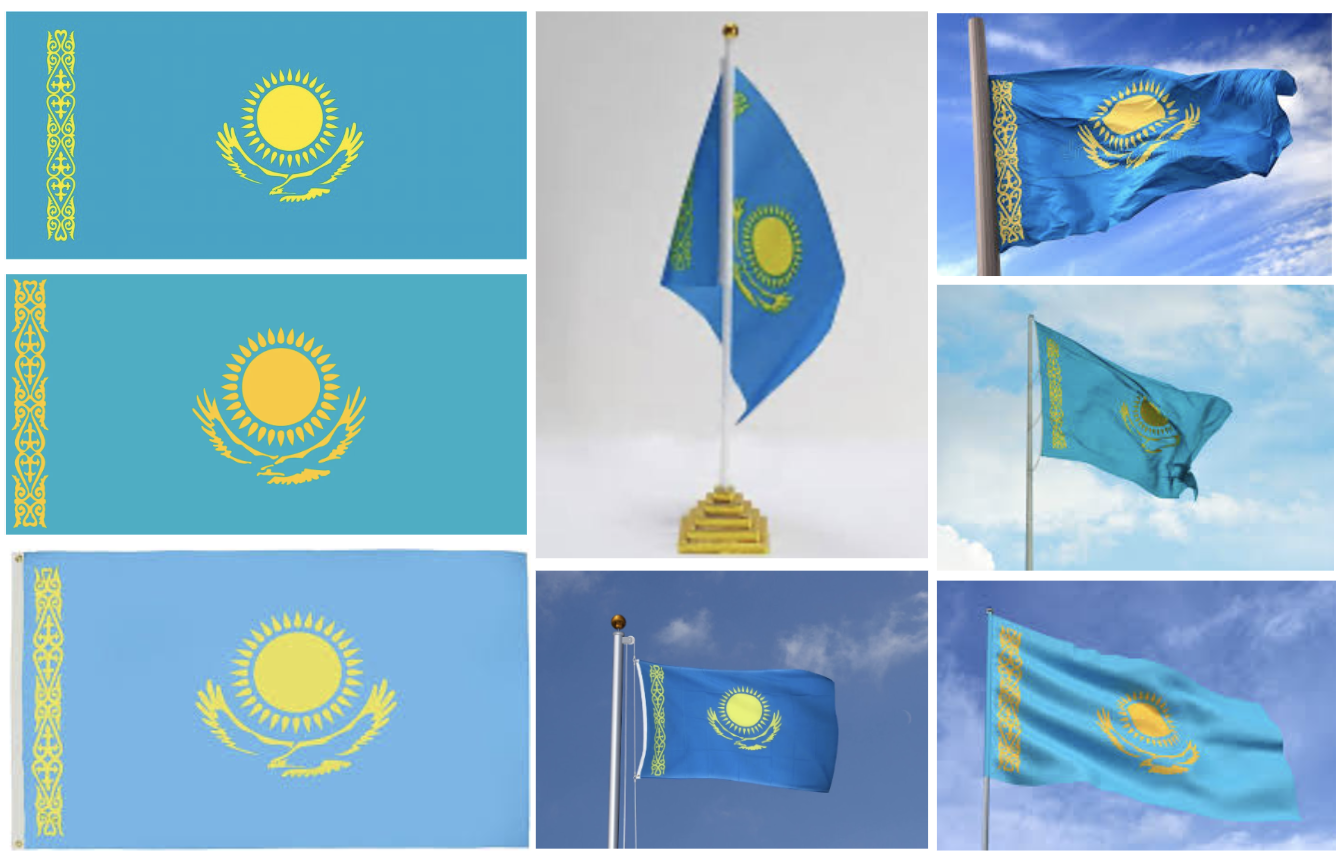}
  
  \caption{Variations in the perceived color of the Kazakhstan flag across different images. The flag appears in shades ranging from light blue to cyan and deep blue, depending on factors such as lighting, printing methods, and linguistic interpretation of color categories.}
  \label{flag}
\end{figure}

Is the flag of Kazakhstan blue, cyan, or turquoise (see Fig. \ref{flag})? Although it is officially described as 'sky blue' \footnote{https://www.akorda.kz/en/state\_symbols/kazakhstan\_flag }, some images depict it as cyan, others as turquoise \footnote{https://www.schemecolor.com/kazakhstan-flag-colors.php}, and depending on display settings, lighting, or cultural context, it may even appear as teal or light blue \footnote{https://www.britannica.com/topic/flag-of-Kazakhstan}. This ambiguity is not only a result of technical color reproduction but also a fundamental issue of linguistic color categorization.

 Language and culture play a crucial role in human color perception, influencing how people perceive, categorize, describe, and interpret colors \cite{Josserand2021Environment, zgen2004Language, He2019Language, Siok2009Language}. For example, in Russian and Kazakh, there are two distinct basic color terms: siniy (dark blue) and goluboy (light blue), whereas in English, both are 'blue'. Such linguistic differences influence how people perceive and classify colors, creating fuzzy category boundaries in human vision. Depending on the observer, the same color may be labeled differently, even though the stimulus remains unchanged. This inconsistency in human color categorization reveals the subjective nature of color perception \cite{Lafer-Sousa2015Striking, Bosten2022Do}. Studies show that aligning human color similarity with large language models can make AI more accurate in reflecting human perception \cite{kawakita2024gromov}. Traditional models, such as RGB and LAB, assign fixed numerical values, failing to account for perceptual and linguistic uncertainty and treating color as a rigid, absolute entity.

To further illustrate this, we examined multiple images of the flag (Fig. \ref{flag}) and noticed slight differences in color, ranging from light blue to cyan. Some of these variations arise from technical factors, such as display settings or printing methods, but others occur due to how color is described and transmitted linguistically. When a color is communicated verbally, such as for printing or design purposes, the chosen term can influence how it is accurately reproduced. What one person calls 'cyan' might be perceived as 'light blue' by another, leading to different interpretations and reproductions. This example illustrates the interaction between language and perception, highlighting the need for a flexible, human-centered approach to color modeling.

% ...
There are specific gaps in existing color models. Current approaches often overlook the inherent uncertainty in color perception, which can be better addressed through the use of fuzzy logic and human-centered experiments. For instance, according to \cite{burambekova2024comparative}, human perception, especially in distinguishing subtle color differences, is not well aligned with traditional models. Another study shows that color constancy, the effect of lighting, angle, and distance on color perception, is not accounted for in these models \cite{da2023color}.

 
While perceptual color models like LAB aim to address uniformity \cite{Kang}, they are still limited in flexibility and adaptability. CIE-based color models, such as CIE LAB and CIE XYZ, were primarily developed for color standardization and reproduction across different displays, printers, and device-independent systems \cite{Fairchild1993, Hasan2012}. While LAB improves upon traditional models like RGB by introducing perceptual uniformity—meaning that a numerical difference in LAB space corresponds to an equal visual difference \cite{Cinko2019Dependence, Sharma2012The}—it does not reflect how humans naturally perceive and categorize colors linguistically which can have imprecise boundaries (e.g., the transition between cyan and blue is not strictly numerical but depends on interpretation).

To address this gap, our study introduces a fuzzy-based color model, COLIBRI, that adapts to human perception and language-driven color distinctions.  COLIBRI enables soft transitions between linguistic categories and offers a more human-aligned color classification that can be applied to various tasks, including image processing. The model is built on fuzzy partitions extracted from extensive human color categorization experiments.


 The COLIBRI model is fundamentally linguistic, built on soft color categorization that aligns with how humans naturally perceive and name colors. Unlike rigid numerical models, COLIBRI uses fuzzy sets to create soft transitions between color categories, accommodating color terms that overlap. The model is particularly inclusive for color-blind individuals, as it provides linguistic descriptions of colors.

The main contributions of this paper are as follows:
\begin{itemize}
    \item \textit{Development of a COLIBRI, a novel human perception-based color model.} The model closely aligns with human color perception by integrating fuzzy logic, addressing perceptual nuances related to hue wideness and saturation dependencies, which traditional models like RGB, HSV, or LAB do not fully capture. To the best of our knowledge, no previous research has documented the construction of fuzzy sets for color attribute specification based on a sample of this size (n=1071).
   
\item \textit{Extraction of fuzzy partitions from large-scale human experiments.} The model identifies and quantifies human-perceived color boundaries through extensive human-centered experiments, creating fuzzy membership functions that reflect real-world perceptual uncertainty and gradual transitions.
    \item \textit{A reproducible experimental framework.} The study presents a standardized setup for human-centered color perception research, ensuring consistency across participants. It employs a multistage experimental design, including linguistic color naming, hue stimuli selection, and human color categorization. Other researchers can adapt and reuse this framework.
    \item \textit{Mechanism for Adaptive Color Categorization}.  The model integrates neural networks to refine fuzzy partitions dynamically, allowing the system to adapt to new datasets or human feedback.
    \item \textit{Analysis of gender differences and color blindness.} The dataset enables the empirical analysis of gender-based perceptual differences and a statistical examination of color-blind individuals' color categorization, ensuring greater inclusivity in color modeling.
    \item \textit{Introducing a new family of color models called Soft Color Models}, which reflect the way people perceive colors with smooth transitions and overlapping categories. Unlike traditional models, they allow a color to belong to multiple categories simultaneously, better aligning with human perception and language.
\end{itemize}

The structure of the paper is as follows. Section II reviews related work, discussing traditional and intelligent color models and identifying gaps in existing research. Section III describes the proposed methodology for developing the COLIBRI model, including the use of fuzzy sets, the HSI color model, and direct rating. Section IV details the multistage experimental design and human color categorization experiments. Section V presents the results and analysis, including fuzzy partitions, model evaluation, and application. Section VI discusses the open questions and challenges and provides a comparison with recent studies. Finally, Section VII concludes the paper with a summary of contributions and future work.


%

\section{Related Work}


\subsection{Color Models}


\begin{table*}[ht]
    \centering
    \caption{Existing fuzzy color spaces: overview}
    \begin{tabular}{|p{2cm}|p{5cm}|p{4cm}|p{5cm}|}
    \hline
\textbf{Fuzzy Color} \newline \textbf{Model} & \textbf{Color Space} & \textbf{Partition creation}& \textbf{Membership function}\\    \hline
        \cite{sugano_colorspace} & HLS is utilized to allow high-density derivation and accurate representation of tone modifiers on a lightness-saturation plane for conical membership functions.  & Geometrically using cones and distance ratios. & Conical membership functions of the Saturation and Lightness. \\ \hline
        \cite{CHAMORROMARTINEZ2007312} & HSI is used due to the semantics of three concepts that are employed by humans to describe colors. & Based on the Munsell color space. & Trapezoidal membership functions of the Hue, Saturation, and Intensity. \\ \hline
        \cite{PakizarShamoi2016} &  HSI color space was utilized due to the perceptual intuitiveness for humans.  & Survey for simplicity, effectiveness, and it requires fewer iterations. & Triangular and Trapezoidal membership functions of the Hue, Saturation, and Intensity.  \\ \hline
        \cite{lammens1995somewhat} & CIE XYZ, CIE L* a* b*, NPP were utilized and compared. CIE L* a* b* had the best result in categorization. & Using a normal Gaussian function where each color category is modeled by focus (the best example) and extent (boundary stimuli). & Normal Gaussian membership function.\\ \hline
        \cite{newdef_fuzzycolorspace} & RGB is used due to its practical relevance by being employed in hardware devices and straightforward Cartesian coordinate representation.  & Derived from the Voronoi cells of the representative colors from the color space. & Linear spline function of the threshold distance that determines the boundaries of the regions in the color space. \\ \hline
        \cite{amante2012fuzzy} & HSV & Defined by Color Module of CSS3 specification. & Trapezoidal membership function Hue, Saturation and Value.\\ \hline
        \cite{shamir2006human} & HSV because it is closer to human perception and more intuitive than RGB. & Partition for Hue is defined by maximum value decribed in \cite{hill1990computer} & Triangular membership function of Hue, Saturation and Value.\\ \hline
        \cite{fuzzycolor_artemotion} & HSI is utilized due to its consistency with human perception and low correlation compared to the RGB color space. & Based on the survey on human color categorization.  & Triangular and Trapezoidal membership functions of the Hue, Saturation, and Intensity.\\ \hline
    \end{tabular}
    \label{fcs_table}
\end{table*}

Color spaces provide a structured way to represent and manipulate colors, each model tailored to specific applications. Among them, the RGB model dominates digital screens, combining red, green, and blue to produce a broad spectrum of colors. However, its inability to separate chromaticity (color) from intensity makes tasks like object recognition and segmentation more complex \cite{García2019Color}. This often necessitates mapping to other color spaces, which is a computationally intensive process. Moreover, RGB is less efficient than other color systems when processing real-world images, as it does not align well with human visual perception \cite{Ibraheem2012, jack2011video}.

To address these limitations, HS* family color models have been developed. The HSV color space, for instance, represents colors through hue, saturation, and brightness, making it more intuitive for designers and artists. Closely related, the HSI model replaces brightness with intensity, allowing for better separation of chromatic information, which simplifies image analysis and segmentation \cite{Zhang2018}.

For applications that require perceptual accuracy, the LAB color model offers a more human-vision-aligned approach. By separating lightness (L*) from color components (a* for green-red and b* for blue-yellow), it ensures that numerical changes correspond more closely to visual differences. This property makes LAB particularly useful for tasks such as color correction and image processing, although its computational complexity can be a drawback \cite{ProLab2021}.

Meanwhile, the CMYK model follows a different logic. Used primarily in printing, it relies on subtractive mixing, where colors are created by filtering out parts of white light. While well-suited for ink-based reproduction, CMYK lacks the perceptual flexibility of models like HSV and LAB, limiting its usefulness in digital analysis \cite{Ibraheem2012, ProLab2021}.

Several studies applied fuzzification to the color spaces mentioned above. Table \ref{fcs_table} presents an overview of various Fuzzy Color Models developed by different authors. A notable trend is the predominant use of HS* color models, largely due to their alignment with human color perception.

According to \cite{book}, color spaces in image processing are classified into device-dependent (specific to a particular hardware), device-independent (ensuring consistent colors across different devices), and perceptual (aligned with human vision to achieve uniform color differences). So, RGB is a device-dependent color model based on additive mixing, commonly used in digital screens but lacking perceptual uniformity \cite{sharma2003digital}. HSV and HSL adjust RGB components for more intuitive color manipulation in design, but do not achieve perceptual uniformity, especially in brightness and saturation handling \cite{hughes2014computer, fairchild2013}. The LAB color model, in contrast, is device-independent and designed for perceptual uniformity, making it useful for color matching despite its computational complexity \cite{chen2012handbook}.

Next, phenomenal color spaces utilize attributes such as hue, saturation, and brightness, and are linear transformations of RGB \cite{ford1998colour}. While widely used in commercial software, such as Adobe's package, phenomenal color spaces have several limitations \cite{charles1999frequently}.
Since they are linear transformations of RGB, they lack chromaticity and white point information. This results in device dependency and introduces a hue discontinuity around 360°, making arithmetic operations in these color spaces more complex \cite{tkalcic2003colour}.

Table \ref{overview_cs} presents the overview of color spaces by category.

Several color models have been derived through direct human experimentation, ensuring alignment with human visual perception. The CIE 1931 XYZ model was based on psychophysical experiments with an initial group of 7 observers, later expanded to 17, who matched colors using controlled mixtures of red, green, and blue light \cite{wyszecki1982}. Similarly, the RGB model was empirically validated through Maxwell’s color-matching experiments \cite{Maxwell1860}. The Munsell Color System was initially developed subjectively but later refined through experiments involving over 40 participants, with further adjustments made in 1943 by the National Bureau of Standards using data from 160 observers \cite{munsell1915, Newhall1943}. In contrast, HSV, HSL, and HSI were designed in the 1970s for digital graphics without direct perceptual studies \cite{Smith1978}, while CMYK, optimized for printing, was developed based on physical ink mixing rather than vision experiments \cite{Yule1967}.


\begin{table*}[ht]
    \small
    \centering
    \caption{Overview of Color Spaces by Category \cite{munsell1915, cie1931, cie1976, sharma2016, fairchild2013, iec1999, adobe1998, fraser2005, itu1994, foley1996, wyszecki1982, pantone2010, aces2014, itu2001}}
    \resizebox{\linewidth}{!}{
    \begin{tabular}{|p{3cm}|p{3.5cm}|p{8.5cm}|}
        \hline
        \textbf{Category} & \textbf{Color Space} & \textbf{Description \& Uses} \\ \hline
        \textbf{Human Perception} & CIE 1931 XYZ & Foundational color model based on human vision; basis for many color standards \cite{cie1931}. \\ \cline{2-3} 
        & CIEUVW & A model adjusting CIE XYZ for improved uniformity. \\ \cline{2-3} 
        & CIELUV, CIELAB & Uniform spaces to represent colors perceptually, common in color matching and graphics \cite{cie1976}. \\ \cline{2-3} 
        & HSLuv & A perceptual variant of HSL, better aligned with human lightness perception \cite{sharma2016}. \\ \cline{2-3} 
        & Munsell Color System & Organized by hue, value, and chroma; foundational for perceptual color categorization \cite{munsell1915}. \\ \hline
        \textbf{Newer Models (CAM)} & CIECAM02 & Modern color appearance model considering viewing conditions and context \cite{fairchild2013}. \\ \hline
        \textbf{RGB Primaries} & sRGB & Standard for digital displays, web colors, and consumer devices \cite{iec1999}. \\ \cline{2-3} 
        & Adobe RGB & Extended gamut for photography and professional imaging \cite{adobe1998}. \\ \cline{2-3} 
        & Adobe Wide Gamut RGB & Extremely wide gamut; suitable for high-quality photo editing. \\ \cline{2-3} 
        & Rec. 2100 & HDR standard for broadcast and streaming, covering a wide color range. \\ \hline
        \textbf{Others with RGB Primaries} & Apple RGB & Used in specific Apple devices, designed to approximate print colors on-screen. \\ \cline{2-3} 
        & ProPhoto RGB & Large gamut, tailored for professional photography and image processing \cite{fraser2005}. \\ \hline
        \textbf{YCbCr and YUV} & YCbCr, YUV & Used in video compression; separates luminance from chrominance for efficient encoding \cite{itu1994}. \\ \hline
        \textbf{Cylindrical Transformations} & HSV, HSL & Used in design applications; separates hue for easier manipulation of color intensity and brightness \cite{foley1996}. \\ \cline{2-3} 
        & LCh & Uniform cylindrical color space, derived from CIELAB, providing improved perceptual alignment. \\ \hline
        \textbf{Subtractive} & CMYK, CMY & Essential for print media, subtractive mixing model using inks and pigments \cite{wyszecki1982}. \\ \hline
        \textbf{Commercial Color Spaces} & Pantone, HKS & Proprietary color systems widely used in branding, design, and color standardization \cite{pantone2010}. \\ \hline
        \textbf{Special-Purpose} & ACES & High dynamic range (HDR) space for film and visual effects, supports high-fidelity color \cite{aces2014}. \\ \hline
        \textbf{Obsolete} & NTSC, PAL, SECAM & Early color standards for analog television, replaced by digital standards \cite{itu2001}. \\ \hline
    \end{tabular}
    }
     \label{overview_cs}
\end{table*}


\subsection{Fuzzy Theory in Color Image Processing}
 

Few studies have successfully integrated fuzzy sets and logic in image processing, including image enhancement, CBIR, image segmentation, etc..

\begin{itemize}
    \item Fuzzy rule-based methods enhance classic image segmentation algorithms by automating the estimation of critical parameters \cite{Karmakar2002}
    \item Type II fuzzy sets can improve image segmentation in non-uniform illumination conditions as they better handle grayness ambiguity \cite{Tizhoosh2005}.
    \item Fuzzy contrast intensification methods based on Gaussian membership functions can be applied for image enhancement tasks \cite{Hanmandlu2003}.
    \item Linguistic labels and their fuzzy representation appear to be beneficial in CBIR systems, where precise matching is not feasible, allowing more flexible queries and enhanced image retrieval \cite{Barranco2005}.
    \item Fuzzy-based image enhancement techniques provide better visualization of the human body (e.g, vascular system), facilitating more accurate medical diagnoses under varying lighting conditions \cite{Gasparri2011}.
    \item Fuzzy Naturals can be used to improve histogram-based image analysis, addressing the limitations of traditional histogram counting methods (e.g, sigma-count method) \cite{Jesus2012}.
\end{itemize}

\subsection{Gaps in Current Research}
There are certain gaps in current computational color models regarding perceptual alignment.

When dealing with human color perception, we cannot overlook the uniformity of the results, which is a primary limitation of existing approaches. The RGB model is not perceptually uniform because a small numerical change in one of the attributes does not correspond to an equivalent change in human-perceived color \cite{Schwarz1987An}. For instance, an analogous distance within the RGB color space may appear different based on the considered color pairs. HS* models are more intuitive for human perception compared to the RGB model. However, they come with certain disadvantages, e.g., the interdependence between I and S components that can complicate color adjustments and the inability to address varying hue wideness \cite{PakizarShamoi2016}. While models like LAB aim to be perceptually uniform, the meaning behind this is that equal distances in the color space should correspond to equal perceived differences \cite{cie1976}. So, they do not capture the soft, language-driven nature of human color perception.

Several perception studies investigate how individuals perceive colors under different lighting conditions and cultural contexts. Specific colors may carry strong symbolic or emotional meanings in some cultures, while being unbiased in others. For example, white is often linked to peace and purity, but it can also symbolize detachment and death in certain cultures \cite{Sruthi2020AN}. 


 In addition to the psychological effects influencing color preferences, environment and conditions \cite{Smithson2005SensoryCA}, context \cite{elliot2012color}, and physical factors in the human brain also affect perception. A notable example is the 'Dress Illusion' phenomenon from 2015, where people saw the same dress as either blue and black or white and gold, depending on how their brains processed the surrounding light and assumed environmental conditions \cite{dress}.

As can be seen from the analysis of existing color model categories, COLIBRI does not fit entirely within any of the established classifications. Perceptual color spaces such as CIE LAB and CIE LUV aim to ensure uniform color differences across devices; their primary goal is color matching rather than reflecting the subjective nature of human color perception. In contrast, COLIBRI introduces the concept of soft color boundaries, where a single color stimulus can belong to multiple hues simultaneously, with varying degrees of membership. COLIBRI accounts for the fact that color perception is inherently fuzzy and is influenced by language, culture, and individual experience, opposing the idea of fixed color boundaries. Therefore, we propose introducing a new category of color models—Soft Color Models—which recognize the gradual transitions between colors and blurred color boundaries. We want to emphasize the distinction between perceptual uniformity and soft categorization by introducing this term to prevent further confusion about the topic.




\section{Methods}
\label{sec:main_section}

\subsection{Proposed Approach}
The COLIBRI model is based on language-driven representation and interpretation of colors. It introduces a soft, human-centered approach, where colors belong to multiple categories with varying degrees of membership, making it more adaptive to real-world perception.

\begin{figure}[tb]
  \centering

 \includegraphics[width=0.5\textwidth]{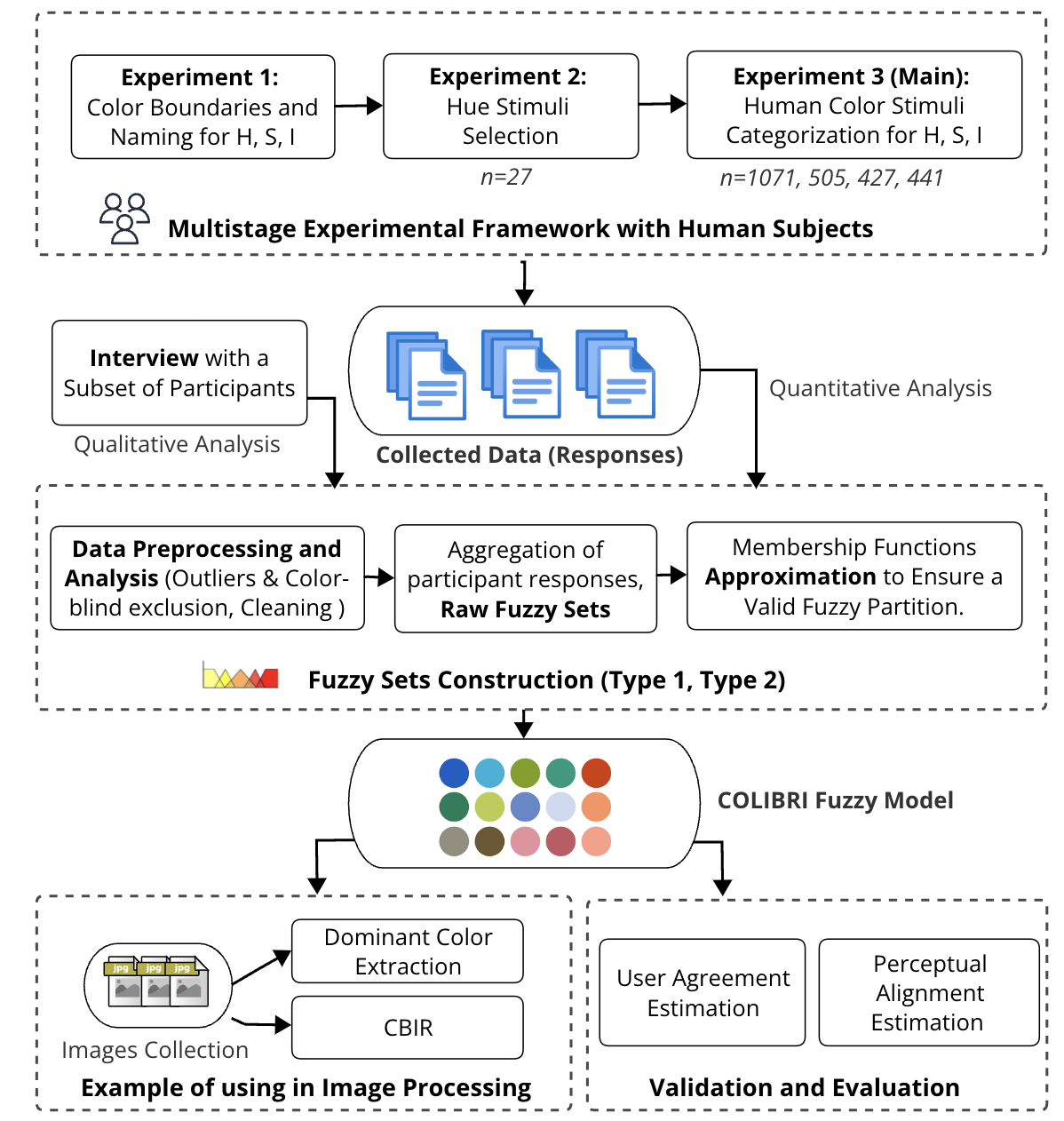}
  
  \caption{High-level methodology of the study, illustrating the multistage experimental framework, fuzzy sets construction, and applications of the COLIBRI fuzzy model.}
  \label{high-level}
\end{figure}

Fig. \ref{high-level} presents the high-level methodology of the study. As can be seen, we begin with a multistage experimental framework for collecting data on human perception of color categorization. The collected data undergoes preprocessing, aggregation, and fuzzy set construction with a valid fuzzy partition through approximation. The resulting COLIBRI fuzzy model is then validated through user agreement estimation and perceptual alignment, with applications demonstrated in image processing, namely, dominant color extraction and content-based image retrieval.

\subsection{Fuzzy Sets and Logic}

Modeling uncertain concepts such as emotions, perception, and colors presents a challenge, as these phenomena lack clearly defined boundaries. Fuzzy set theory, first introduced by Lotfi Zadeh \cite{ZADEH1965338}, provides a framework for representing such concepts by allowing elements to have a membership degree ranging from 0 to 1, rather than the binary classifications of classical set theory. This flexibility makes fuzzy sets particularly effective in modeling uncertainty and gradual transitions in human perception.


Fuzzy sets are useful in modeling colors based on how people perceive them, as they can handle uncertainty and ambiguity effectively. They assign a degree of membership to each element, allowing for a more detailed understanding of uncertainty. Additionally, experts can help create fuzzy partitions that reflect people's subjective experiences with color \cite{PakizarShamoi2016}.

\subsubsection{Fuzzy Sets and Membership Functions}

A fuzzy set, introduced by Zadeh \cite{ZADEH1965338}, is a mathematical tool for representing uncertainty and imprecision by allowing elements to have degrees of membership in the range \([0,1]\) rather than binary classifications. Formally, a fuzzy set \( A \) in a universe of discourse \( X \) is defined by a membership function:
\begin{equation}
\mu_A: X \to [0,1],
\end{equation}
where \( \mu_A(x) \) denotes the degree to which an element \( x \in X \) belongs to the fuzzy set \( A \). This flexibility makes fuzzy sets particularly useful for modeling vague and subjective concepts such as human perception and decision-making.



Fuzzy sets are widely used in tasks related to fuzzy color representation and processing. They have been extensively applied in defining fuzzy color spaces \cite{fuzzycolorspaces, newdef_fuzzycolorspace, granular_fuzzycolorcat, imagebased_fuzzycolorspace, fuzzycolor_artemotion, PakizarShamoi2016}, where they enable flexible and perceptually relevant categorizations of colors based on human vision. Additionally, fuzzy logic plays a crucial role in fuzzy color image retrieval \cite{fuzzy_colorimageret, fuzzy_colorhisto_imageret, fcth_imageret, fuzzy_hcp_ecom}, where it enhances retrieval accuracy by handling color variations and uncertainties in image databases.

\paragraph{Type-2 Fuzzy Set}
Type-2 fuzzy sets, introduced by Zadeh \cite{ZADEH1975}, extend type-1 fuzzy sets by incorporating uncertainty in their membership functions. Instead of assigning a single membership degree \( \mu_A(x) \in [0,1] \), type-2 fuzzy sets define membership degrees as fuzzy sets themselves, allowing them to model higher levels of imprecision.
A type-2 fuzzy set \( \tilde{A} \) in the universe \( X \) is defined as:  
\begin{equation}
\tilde{A} = \left\{ (x, \mu_{\tilde{A}}(x)) \mid x \in X \right\},
\end{equation}
where \( \mu_{\tilde{A}}(x) \) is a type-1 fuzzy set in \([0,1]\) with a secondary membership function \( \mu_{\tilde{A}}(x, u) \). The \textit{footprint of uncertainty (FOU)} is the region between the \textit{upper} and \textit{lower membership functions}, forming an \textit{interval type-2 fuzzy set (IT2FS)}:
\begin{equation}
\bar{\mu}_{\tilde{A}}(x) = \sup_{u} \mu_{\tilde{A}}(x, u), \quad
\underline{\mu}_{\tilde{A}}(x) = \inf_{u} \mu_{\tilde{A}}(x, u).
\end{equation}
This additional uncertainty handling makes type-2 fuzzy sets particularly effective for applications involving human perception and decision-making.

\paragraph{Intuitionistic Fuzzy Sets}
An intuitionistic fuzzy set (IFS) extends traditional fuzzy set theory by incorporating not only the degree of membership $\mu$ but also the degree of non-membership $\nu$ for each element of a set, such that $0 \leq \mu(x) + \nu(x) \leq 1.$  The remaining part $\pi_A(x) = 1 - \mu_A(x) - \nu_A(x) \quad$ captures the degree of uncertainty or indeterminacy. This allows a more nuanced representation of uncertainty. 

IFS are widely implemented in tasks of image processing such as segmentation \cite{inst_medimage_seg, inst_leukocytes_seg, inst_multspatial_seg, inst_image_segment, inst_med_edge_enhan}, similarity measure for image retrieval \cite{inst_afsari2010, inst_afsari2014fuzzy, inst_xu2012similarity}, and image enhancement \cite{inst_med_edge_enhan, inst_image_enhancement, inst_contrast_enhancement}.

\cite{inst_afsari2010} proposed an image retrieval system that extracts images based on the dominant color in the HSV color space to get features based on eight fundamental colors of hue and eight levels of value, which are then joined into two-dimensional intuitionistic fuzzy sets. Similarity is calculated using a fuzzy quantity between two IFS images. \cite{inst_leukocytes_seg} applied IFS to segment leukocytes in color images by modeling uncertainty, using the hue component in HSV color space to capture pixel ambiguity. This approach preserves information more effectively than grayscale conversion and enhances segmentation accuracy by taking into account hesitation in pixel classification. 

\paragraph{Types of membership functions}

\begin{figure*}[tb]
    \centering
    \begin{subfigure}[b]{0.3\textwidth}
        \centering
        \includegraphics[width=\textwidth]{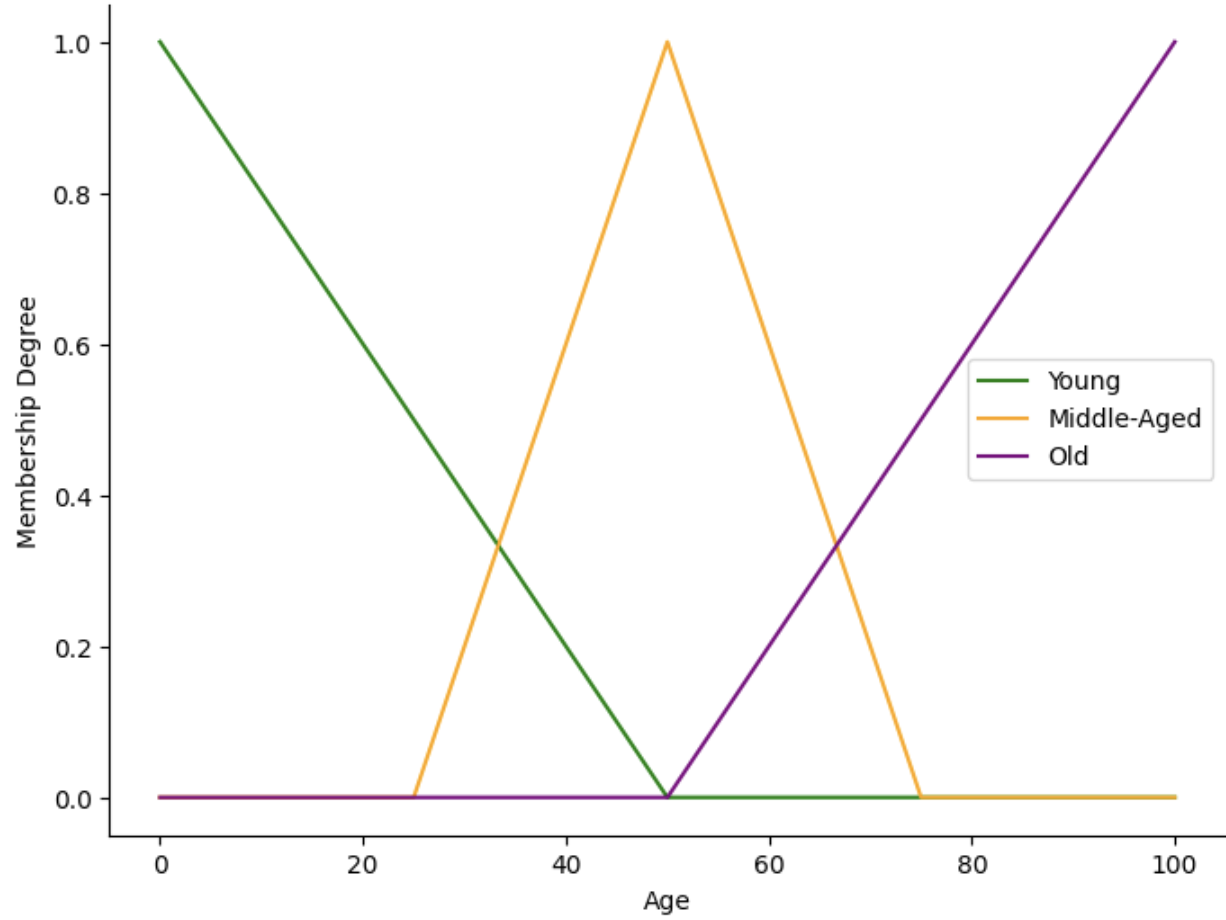}
        \caption{Triangular membership function}
        \label{fig:triangular}
    \end{subfigure}
    \hfill
    \begin{subfigure}[b]{0.3\textwidth}
        \centering
        \includegraphics[width=\textwidth]{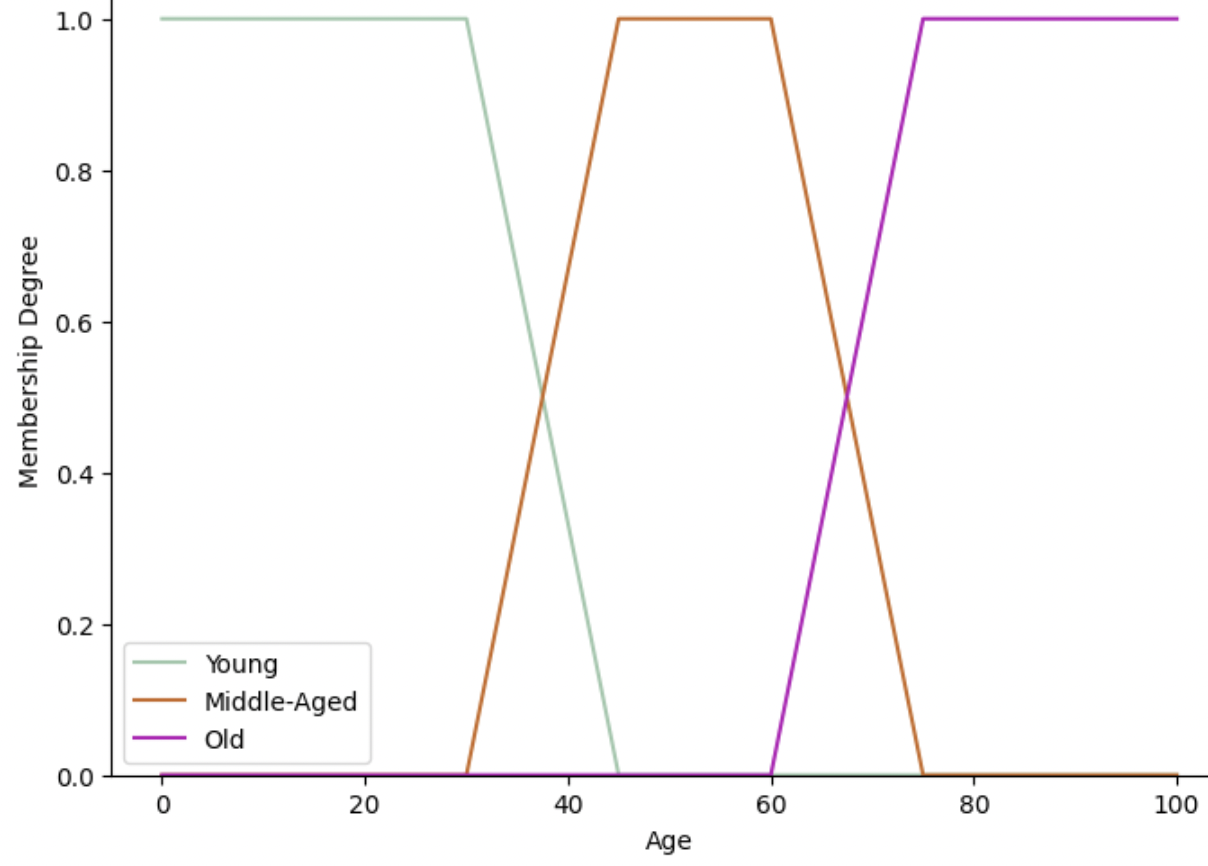}
        \caption{Trapezoidal membership function}
        \label{fig:trapezoidal}
    \end{subfigure}
    \hfill
    \begin{subfigure}[b]{0.3\textwidth}
        \centering
        \includegraphics[width=\textwidth]{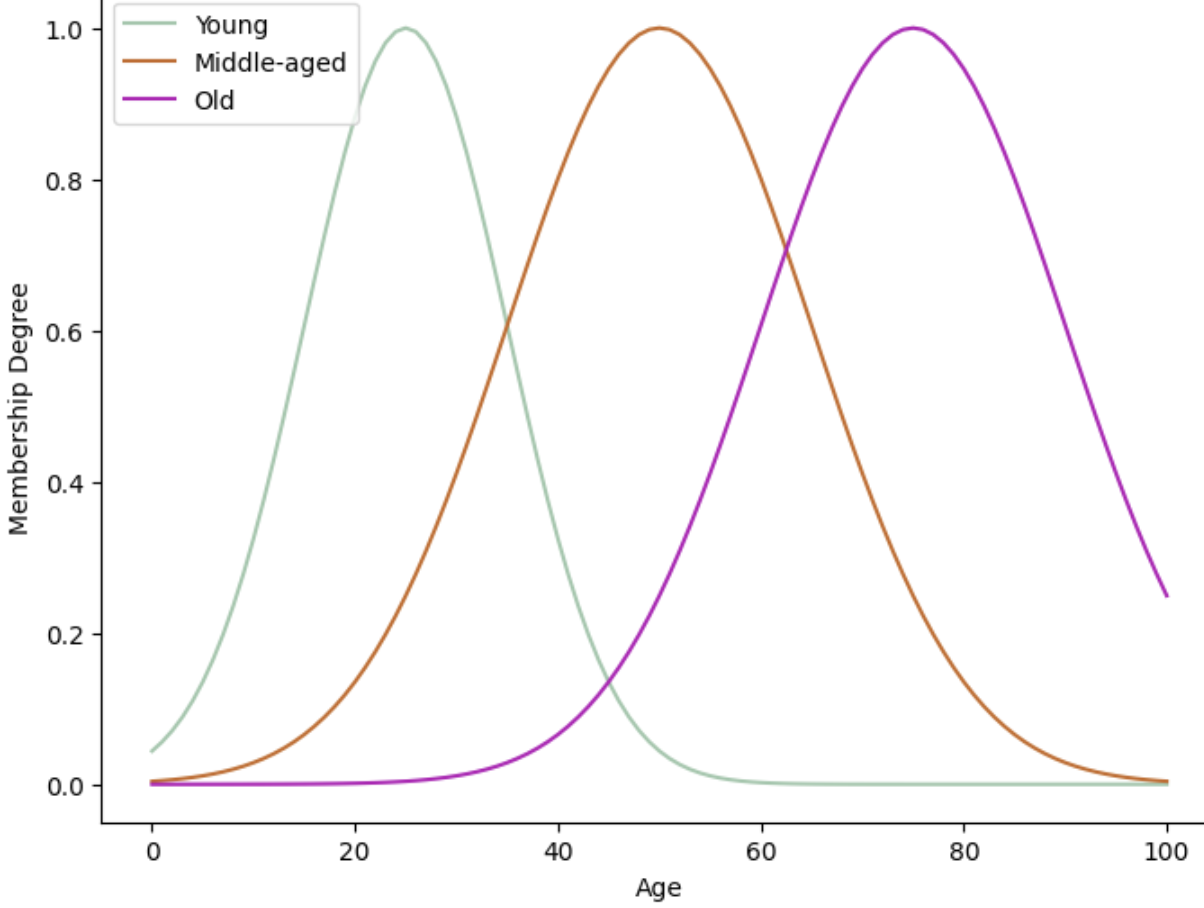}
        \caption{Gaussian membership function}
        \label{fig:gaussian}
    \end{subfigure}

    \caption{Comparison of different fuzzy membership functions}
    \label{fig:membership_functions}
\end{figure*}

Figure \ref{fig:membership_functions} illustrates standard fuzzy membership functions, including triangular, trapezoidal, and Gaussian, each used to represent uncertainty and gradual transitions. 

\textbf{Triangular membership function} \\
The triangular membership function is a simple but widely used function in fuzzy logic systems, characterized by its linear rise and fall. It is defined by three parameters: $a$, $b$, and $c$, where $a$ and $c$ are the lower and upper base points, and $b$ is the peak point of the triangle. The triangular membership function is especially useful when the transition between sets needs to be approximated with a single peak and linearly decreasing sides. Its mathematical form is as follows:
\begin{equation}
 \mu(x) = \begin{cases}
    0, & x \leq a \\
    \frac{{x - a}}{{b - a}}, & a \leq x \leq b \\
    \frac{{c - x}}{{c - b}}, & b \leq x \leq c \\
    0, & x \geq c
\end{cases} 
\label{triangular_mf}
\end{equation}

\
\textbf{Trapezoidal membership function} \\
The trapezoidal membership function is one of the standard shapes used in fuzzy logic systems due to its flexibility and simplicity. It is defined by four points, creating a shape that resembles a trapezoid. The parameters $a$ and $d$ control the trapezoid's left and right base points, respectively, and the parameters $b$ and $c$ control the trapezoid's left and right top points, respectively.
The general form of a trapezoidal membership function $\mu(x)$ can be described mathematically as:
\begin{equation}
     \mu(x) = \begin{cases}
        0, & x \leq a \\
        \frac{{x - a}}{{b - a}}, & a \leq x \leq b \\
        1, & b \leq x \leq c \\
        \frac{{d - x}}{{d - c}}, & c \leq x \leq d \\
        0, & x \geq d
    \end{cases} 
\label{trapezoidal_mf}
\end{equation}

\textbf{Gaussian membership function} \\
The Gaussian membership function is another standard fuzzy set membership function, characterized by its smooth and symmetric bell-shaped curve. It is defined by two parameters: the center of the peak $c$ and the standard deviation $\sigma$, which determines the spread of the function. The Gaussian membership function is handy when smooth transitions between membership values are required in fuzzy systems. 
The mathematical representation of the Gaussian membership function is:
\begin{equation}
    \mu(x) = \exp\left(-\frac{(x - c)^2}{2\sigma^2}\right)
\label{gaussian_mf}
\end{equation}

\textbf{Sigmoidal membership function} 
The sigmoidal membership function is characterized by its smooth, S-shaped curve, which can either increase or decrease depending on its parameters. It is defined by two parameters: the slope $a$, which controls the steepness of the curve, and the midpoint $c$, which determines the center of the transition from 0 to 1 (or from 1 to 0, in the case of a decreasing sigmoid). 
The general form of the sigmoidal membership function is given by:
\begin{equation}
    \mu(x) = \frac{1}{1 + e^{-a(x - c)}}
\label{sigmoidal_mf}
\end{equation}

\subsubsection{Linguistic Variables}

The concept of \textit{linguistic variables} was introduced by Zadeh \cite{ZADEH1975} as an extension of classical variables to represent imprecise or qualitative information. Unlike numerical variables, a linguistic variable takes words or phrases as values, which are described by fuzzy sets. Formally, a linguistic variable \( V \) is defined as:
\begin{equation}
V = (X, T(X), U, G, M),
\end{equation}
where:
\begin{itemize}
    \item \( X \) is the name of the variable (e.g., ``Temperature"),
    \item \( T(X) \) is the set of possible linguistic terms (e.g., \{``Cold", ``Warm", ``Hot"\}),
    \item \( U \) is the universe of discourse (e.g., numerical temperature values),
    \item \( G \) is the syntactic rule for generating linguistic terms,
    \item \( M \) is a semantic rule that maps each term in \( T(X) \) to a fuzzy set in \( U \).
\end{itemize}

\subsubsection{Fuzzy Rules}

Fuzzy rules are conditional statements in the form of "IF-THEN" expressions that define relationships between input and output variables in a fuzzy system \cite{Zadeh1988}. These rules use linguistic terms and fuzzy logic operators to model complex, imprecise, or uncertain phenomena. By combining multiple fuzzy rules, a system can approximate human reasoning and make flexible, interpretable decisions based on gradual transitions rather than strict binary classifications. For example: 
\begin{align*}
\text{Rule 1: } & \text{IF } H \text{ is around 0°} \text{ AND } S \text{ is High} \\
                & \text{THEN } \text{Color is Red}. \\
\text{Rule 2: } & \text{IF } H \text{ is around 120°} \text{ AND } S \text{ is Medium} \\
                & \text{THEN } \text{Color is Green}. \\
\text{Rule 3: } & \text{IF } H \text{ is around 240°} \text{ AND } S \text{ is Low} \\
                & \text{THEN } \text{Color is Blue}.    
\end{align*}
These fuzzy rules illustrate how fuzzy logic is applied to color classification. Instead of strict boundaries, colors are defined using linguistic variables, allowing for gradual transitions and handling perceptual uncertainty.

\subsection{HSI Color Model}



As a base for the study experiments, the HSI color space will be used. 

HSI color space represents colors by the three attributes: hue, saturation, and intensity \cite{fuzzycolor_artemotion}, providing an intuitive semantics of color names \cite{Shamoi2015DeepCS}. Despite the HSI color space's alignment with human perception, most digital devices and image processing systems primarily use the RGB color space. Therefore, it is necessary to implement mechanisms for converting between HSI and RGB color spaces. This ensures compatibility with device standards while retaining the perceptual advantages of HSI representation.

\begin{figure}[tb]
 \includegraphics[width=0.45\textwidth]{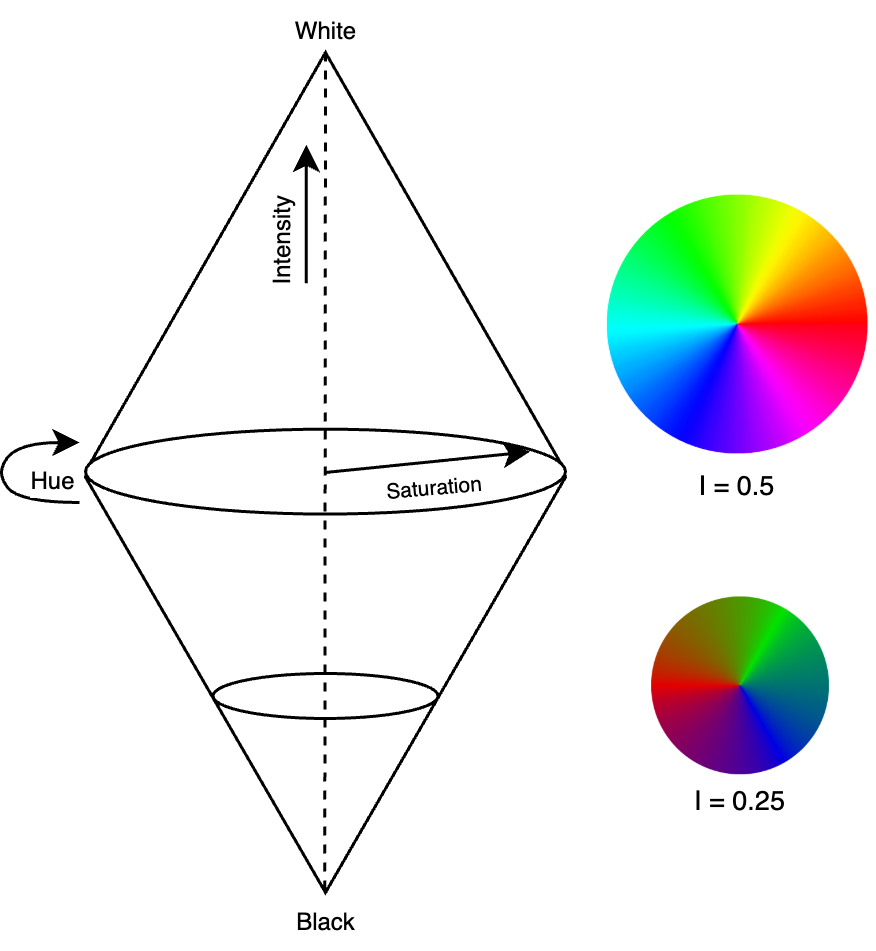}
 \caption{HSI color model}
 \label{fig:hsi}
\end{figure}

A potential concern in the RGB-to-HSI conversion process is the occurrence of many-to-one mappings, where multiple RGB values can correspond to the exact HSI representation or vice versa. For example, fully desaturated colors (grayscale) are represented in HSI with S=0, making the hue (H) irrelevant and causing different RGB values to map to identical HSI representations \cite{kamiyama2021color, zhi2020fpga}.


    

The following equations give the conversion from HSI to RGB for each sector \cite{Taguchi1970}: 

\begin{itemize}
\item If \( H \in [0^\circ, 120^\circ] \):
\begin{equation}
    b = I \times (1 - S)
\end{equation}
\begin{equation}
  r = I \times \left( 1 + \frac{S \cdot \cos H}{\cos(60^\circ - H)} \right)  
\end{equation}
\begin{equation}
    g = 3 \times I - (r + b)
\end{equation}
\item If \( H \in (120^\circ, 240^\circ] \):
\begin{equation}
    H = H - 120
\end{equation}
\begin{equation}
    r = I \times (1 - S)
\end{equation}
\begin{equation}
    g = I \times \left( 1 + \frac{S \cdot \cos H}{\cos(60^\circ - H)} \right)
\end{equation}
\begin{equation}
    b = 3 \times I - (r + g)
\end{equation}
\item If \( H \in (240^\circ, 360^\circ] \):
\begin{equation}
    H = H - 240
\end{equation}
\begin{equation}
    g = I \times (1 - S)
\end{equation}
\begin{equation}
    b = I \times \left( 1 + \frac{S \cdot \cos H}{\cos(60^\circ - H)} \right)
\end{equation}
\begin{equation}
    r = I \times 3 - (g + b)
\end{equation}

\end{itemize}

\noindent
where:
\begin{itemize}
    \item \( H \) — hue angle in degrees, \( H \in [0^\circ,360^\circ] \)
    \item \( S \) — saturation, \( S \in [0,1] \),
    \item \( I \) — intensity, \( I \in [0,1] \).
\end{itemize}

\subsection{Ishihara Test for Color Blindness}
Color vision plays a crucial role in daily life, from basic things like choosing clothes or food to identifying traffic lights. However, a lot of people struggle with color vision deficiency (CVD), commonly known as color blindness, which affects various aspects of daily life. Individuals with CVD face challenges in activities that require color discrimination, which can affect their quality of life and career opportunities \cite{Machado2009A} \cite{Zhu2021Image}.

In 1917, Dr. Shinobu Ishihara introduced the method for detecting color vision deficiencies (CVD)\cite{ishihara1918tests}. This test is still one of the most widely used instruments. It consists of a series of plates, each containing a pattern of colored dots that form numbers or pathways visible to those with normal color vision, but difficult or impossible to recognize for individuals with color blindness. The simplicity and effectiveness of the method have made it a standard in medical and occupational assessments.

\begin{figure}[tb]
 \includegraphics[width=0.45\textwidth]{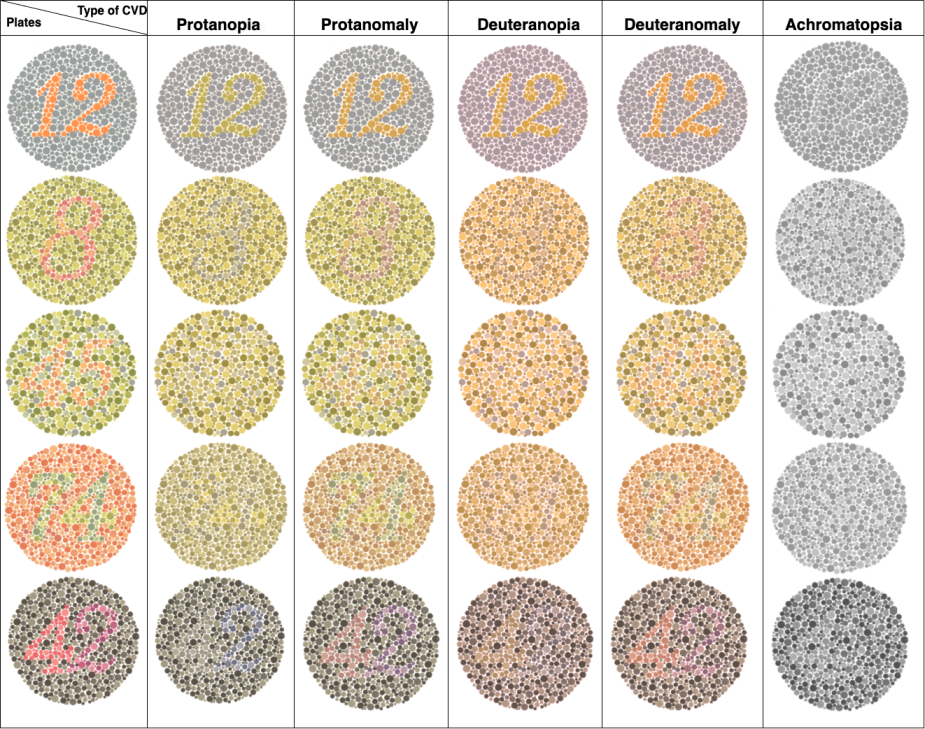}
 \caption{Ishihara color model}
 \label{fig:ishihara}
\end{figure}

Color blindness, or color vision deficiency (CVD), occurs when the cone cells in the retina do not function properly. The main types include (see Fig. \ref{fig:ishihara}):
\begin{itemize}
    \item Protanopia - A form of red-green color blindness caused by the absence of red (long-wavelength) cone cells, making it difficult to distinguish between red and green hues \cite{Roskoski2017Guidelines}
    \item Deuteranopia - Another type of red-green color blindness, resulting from defective green (medium-wavelength) cone cells. Like protanopia, it impairs the ability to differentiate between red and green shades \cite{Roskoski2017Guidelines}
    \item Achromatopsia - A rare condition characterized by the complete absence of color vision. Individuals with achromatopsia perceive the world in shades of gray and often experience light sensitivity and reduced visual acuity.
\end{itemize}


\subsection{Direct Rating and Color Categorization}
  The direct rating (DR) method is a technique where participants assign numerical or categorical values to specific attributes of an object or stimulus based on their subjective perception \cite{bottom2000}. This method is widely used in psychology, marketing, social sciences, and human-computer interaction to quantify subjective judgments and variations.  Research has shown that DR is more reliable and accurate in test-retest situations and objectively verifiable perceptual tasks \cite{bottom2000, bottom2013}. DR also produces more favorable reactions from subjects and leads to more consistent decision-making \cite{bottom2000}.

 In a typical DR, human subjects are presented with a stimulus (e.g., an image, sound, product, or concept). Based on their perception or preference, they provide a rating on a predefined scale (e.g., 1 to 5) or choose a category. The collected ratings are then aggregated and analyzed to determine patterns or classifications.


Several studies employ direct rating to measure color preferences and categorize color stimuli \cite{Pastoor1990Legibility, Huang2012The, 7044882, Shamoi2020, fuzzy_hcp_ecom}. This study uses the DR method to classify color stimuli based on human perception. Specifically, participants were presented with a set of color samples and asked to categorize each stimulus into a color category (e.g., orange or red) based on their perception. Since human perception of color is not rigid, a single color stimulus could be associated with multiple linguistic labels, each with a different degree of membership.

The percentage of participants who selected a given label for a color stimulus was used to compute its membership function within the corresponding fuzzy set. This allows for a soft classification, where a stimulus may belong to multiple categories with varying degrees.

Let us examine how the results of direct rating-based experiments can be used to create fuzzy sets, using a \textit{Hue} fuzzy variable as an example.

Let \( L = \{l_1, l_2, ..., l_m\} \) be the set of linguistic labels (e.g., \textit{red, orange, yellow}), \( S = \{s_1, s_2, ..., s_n\} \) be the set of color stimuli presented to participants. Each linguistic term is modeled using a fuzzy MF:

\begin{equation}
\mu_{l_j}(s) : S \to [0,1]
\end{equation}

where:
\begin{itemize}
    \item \( s \) is the stimulus (color value, e.g., hue angle).
    \item \( \mu_{l_j}(s) \) is the membership degree of \( s \) in fuzzy set \( l_j \).
\end{itemize}

Each stimulus \( s_i \) is classified by participants into multiple labels \( l_j \), leading to the initial, or raw membership calculation:

\begin{equation}
\tilde{\mu}_{l_j}(s_i) = \frac{N_{i,j}}{N}
\end{equation}

where:
\begin{itemize}
    \item \( \tilde{\mu}_{l_j}(s_i) \) is the raw membership value of \( s_i \) in label \( l_j \).
    \item \( N_{i,j} \) is the number of participants who classified \( s_i \) as \( l_j \).
    \item \( N \) is the total number of participants evaluating \( s_i \).
\end{itemize}

At any given point \( s \), the sum of membership degrees across all hue categories must satisfy the fuzzy partition constraint:


\begin{equation}
\sum_{j=1}^{m} \mu_{l_j}(s_i) = 1, \quad \forall s_i \in S.
\end{equation}

This ensures a valid fuzzy partition where no region is left undefined and no overestimation occurs.

Each linguistic term is modeled using either a triangular or trapezoidal MF. Final fuzzy sets need to be obtained by limiting the shape of raw fuzzy sets to triangular or trapezoidal ones. We also need to ensure smooth transitions between neighboring fuzzy sets representing the hues, just as in real life, e.g., the gradual transition of rainbow colors into one another.  So, e.g., for the triangular MF having $a,b,c$ parameters, $a$ is the left boundary (shared with the previous fuzzy set), $b$ is the peak (center of the color category), $c$ is the right boundary (shared with the next fuzzy set).

Approximation is required to adjust the MFs mathematically while ensuring that the fuzzy partition constraint is satisfied. Since each fuzzy set overlaps with its neighbors, the parameters are adjusted so that at every \( s \), the sum is exactly 1.


We use the following parameter approximation strategy:
\begin{itemize}
    \item Shared boundaries, the end of one fuzzy set is the start of the next, so there are no gaps between fuzzy sets:  
    \begin{equation}
    c_j = a_{j+1}
    \end{equation}
    \item Peak alignment, i.e., the peak of a neighboring fuzzy set must be positioned so that the sum remains 1:
    \begin{equation}
    \mu_{l_j}(b_j) + \mu_{l_{j+1}}(b_j) = 1.
    \end{equation}
    \item Rescaling overlapping regions to enforce strict fuzzy partitioning. If at any \( s \), the sum exceeds 1:
    \begin{equation}
    \mu'_{l_j}(s) = \frac{\mu_{l_j}(s)}{\sum_{k=1}^{m} \mu_{l_k}(s)}
    \end{equation}
    This rescales all membership values proportionally.
\end{itemize}


This strategy ensures smooth transitions between color categories and valid fuzzy membership values (sum is 1 at all points).


 \subsection{Outlier Identification}


In our study on color categorization involving over 1,000 participants, identifying outliers was essential to ensure the accuracy and reliability of the data. Given the subjective nature of the responses, where participants evaluated and categorised colors, outlier management was particularly critical. As Sullivan et al. point out, choosing the appropriate method for assessing outliers is essential to accurately reflect the study's social experiment context and ensure the validity of its conclusions \cite{sullivan2021so}. Rosenthal also highlights the necessity of adapting statistical techniques to effectively handle data influenced by human behaviors and perceptions \cite{rosenthal2011statistics}.
The Q3 + 1.5 IQR method is particularly valued for its ability to detect extreme values by considering the spread and central tendency of the data, according to Tukey \cite{tukey1977exploratory}. 
The 1st and 99th percentile methods are employed to address the extremes in the dataset, capturing the most significant deviations from general responses, as Pearson describes in his study on outliers in process modeling and identification \cite{arimie2020outlier}.
These methods establish an approach to preprocessing that significantly enhances the reliability and validity of our research findings, which is essential for advancing our understanding of human color categorization.

 
 \subsection{Ethics statement}
 The Ethics Committee of Kazakh-British Technical University approved this research project, which involved data collection and multiple experiments with human subjects. An open invitation to participate was distributed via email, social media, and institutional messaging platforms, reaching a broad audience. Participation took place exclusively offline to ensure consistency by using the same computers under identical settings for all participants.

Prior to the study, participants were given a written document outlining the research purpose, experimental procedure, and their rights. This information was provided in Kazakh, Russian, and English to accommodate individuals with diverse linguistic backgrounds. A video presentation explaining the procedure was also shown, and participants were encouraged to ask questions for further clarification.

It was emphasized that the study involved minimal risk and that participants had the freedom to withdraw at any time without consequences. Each participant provided signed informed consent before beginning the experiment. The study was conducted under the supervision of designated observers to ensure adherence to ethical guidelines. No personal identifying information was collected, and all data were handled in a confidential manner.

\section{Design of Experiments}

\begin{figure*}[tb]
  \includegraphics[width=1\textwidth]{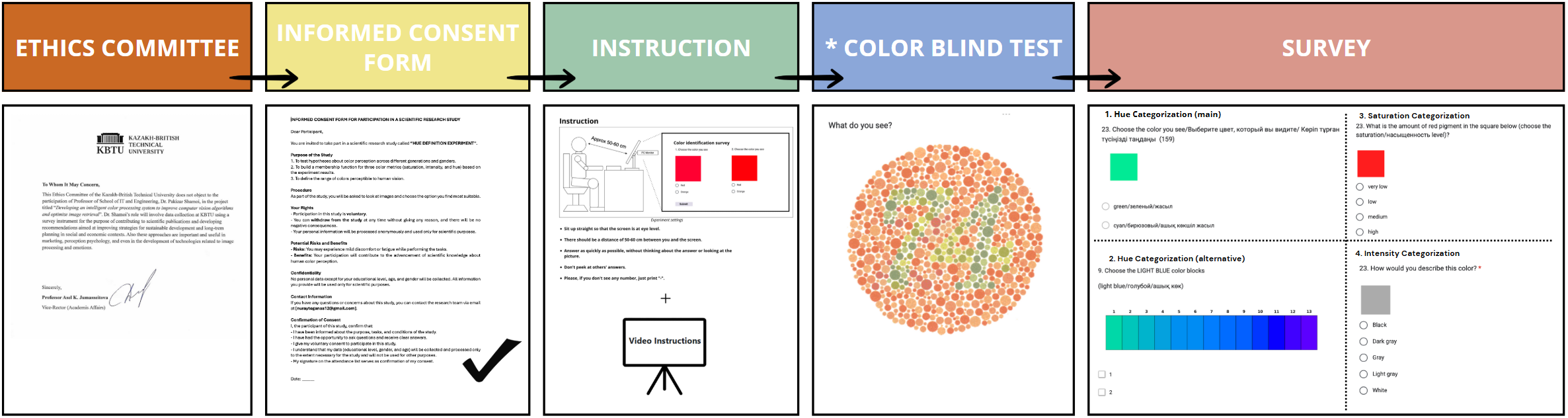}
  \caption{Experimental pipeline/protocol. We conducted an experiment with participants after obtaining approval from the ethics committee and ensuring that they had signed an approved informed consent form before the survey. Subsequently, instructions were provided, and to ensure clear results, a color-blind test was administered before the hue categorization surveys. }
  \label{figure_SE}
 \end{figure*}

\subsection{Experiment Settings}
\label{subsec:Experiment Settings}

\begin{table*}[tb]
  \caption{Settings used in the multistage experimental framework}
    \centering
    \renewcommand{\arraystretch}{1.2}
    \begin{tabular}{|p{1.3cm}|p{1.3cm}|p{1cm}|p{5.7cm}|p{3cm}|p{1.5cm}|}
        \hline
        \textbf{Parameter} & \textbf{Color Presentation Type} & \textbf{Number of Color Stimuli} & \textbf{Description/Task} & \textbf{Results} & \textbf{Number of participants (gender ratio (\%))}\\
        \hline
        \multicolumn{6}{|c|}{\textbf{Experiment 1: Perceptual Color Boundaries and Naming}} \\
        \hline
        Hue        & Composite & 360 & Participants were shown a continuous color spectrum and asked to identify how many distinct hue categories they perceived. & Nine linguistic hue categories: Red, Orange, Yellow, Green, Cyan, Light Blue, Blue, Violet, and Magenta. & 7 (71.4\% F \& 28.6\% M) \\
        \hline
        Saturation & Composite & 100 & Participants were presented with a red saturation gradient ranging from desaturated (white) red to fully saturated red and asked to identify the number of distinct saturation levels they perceived. & Four saturation levels: Very low, Low, Medium, High & 7 (71.4\% F \& 28.6\% M) \\
        \hline
        Intensity  & Composite & 255 & Participants were presented with a grayscale gradient ranging from black to white and asked to identify the number of distinct intensity (brightness) levels they perceived. & Five intensity categories: Black, Dark gray, Gray, Light gray, White & 7 (71.4\% F \& 28.6\% M) \\
        \hline
        \multicolumn{6}{|c|}{\textbf{Experiment 2: Hue Stimuli Selection Experiment}} \\
        \hline
        Hue & Parallel & 360 & The 360-degree hue spectrum was initially divided into hue steps of 3 degrees. & 120 color stimuli & 7 (71.4\% F \& 28.6\% M) \\
        \hline
        Hue & Parallel & 120 & Participants were presented with a series of color stimuli containing hues within predefined boundaries of nine hue categories and were asked to indicate how many distinct colors they perceived within each category. & Final 45 color stimuli & 27 {(37\% F \& 63\% M)} \\
        \hline
        \multicolumn{6}{|c|}{\textbf{Experiment 3 (Main): Human Color Stimuli Categorization}} \\
        \hline
        Hue (main) & Single & 45 & Participants were shown isolated color stimuli and asked to choose between two options representing different hues. & Value for constructing the membership function & 1071 (40.3\% F \& 59.7\% M)\\
        \hline
        Hue (alternative) & Parallel & 45 & Participants were shown a series of colored blocks and asked to choose which blocks seemed a certain color to them, depending on the color that was presented. & Value for constructing the membership function & 505 (47.1\% F \& 52.9\% M) \\
        \hline
        Saturation & Single & 26 & Participants identified the saturation level according to their perception and classified the color stimuli into categories. & Value for constructing the membership function & 427 (50.11\% F \& 49.89\% M) \\
        \hline
        Intensity  & Single & 29 & Participants identified the intensity level according to their perception and classified the achromatic color stimuli into categories. & Value for constructing the membership function & 441 (49.89\% F \& 50.11\% M) \\
        \hline
    \end{tabular}
    \label{tab:exp_conditions}
\end{table*}

 \begin{figure}[tb]

 \includegraphics[width=0.5\textwidth]{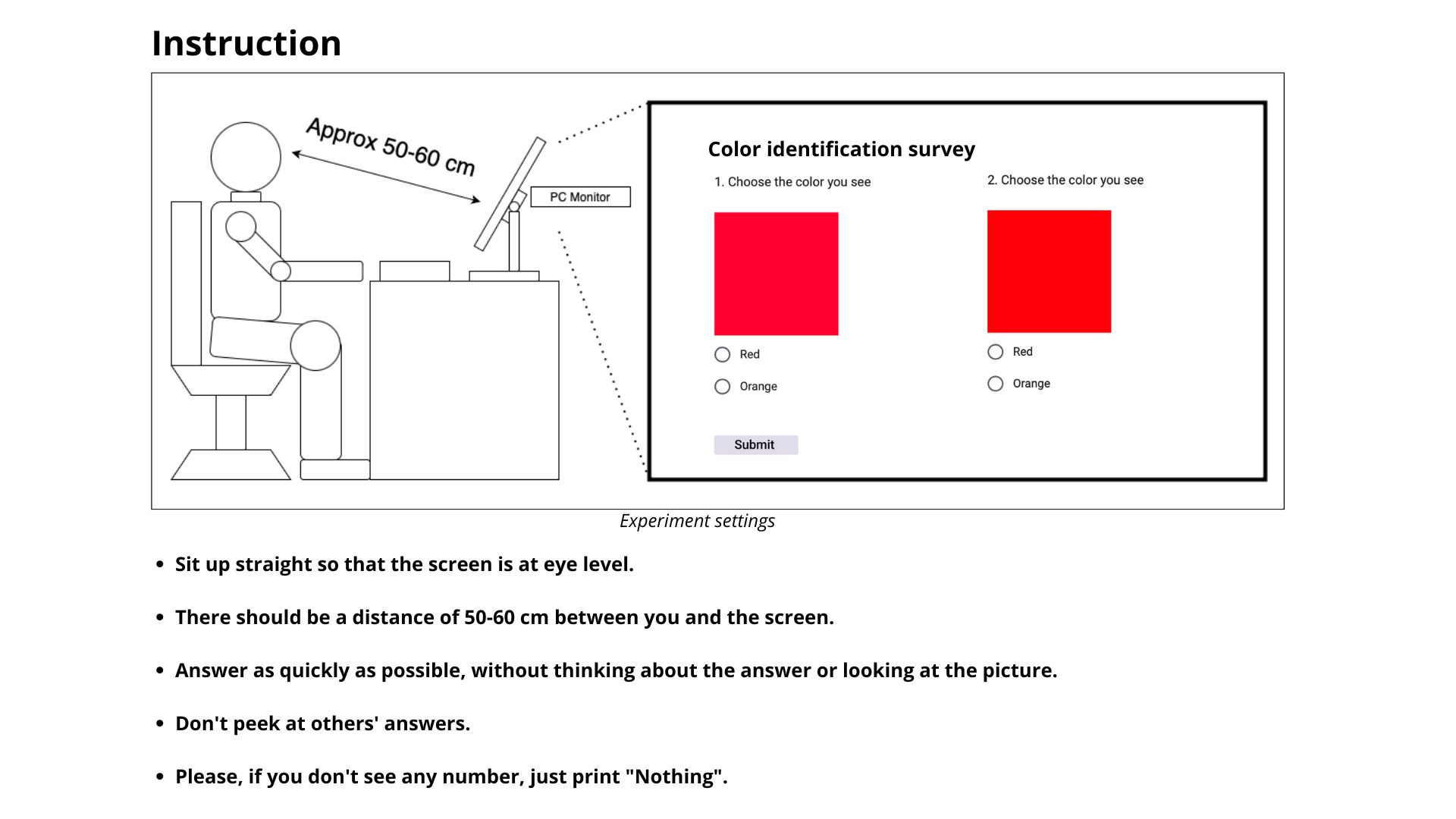}
  
  \caption{
  Instructions for participants
 }
  \label{fig:ins_1}
\end{figure}

\begin{figure*}[h]
  \centering
  \makebox[\textwidth][c]{
 \includegraphics[width=0.85\textwidth]{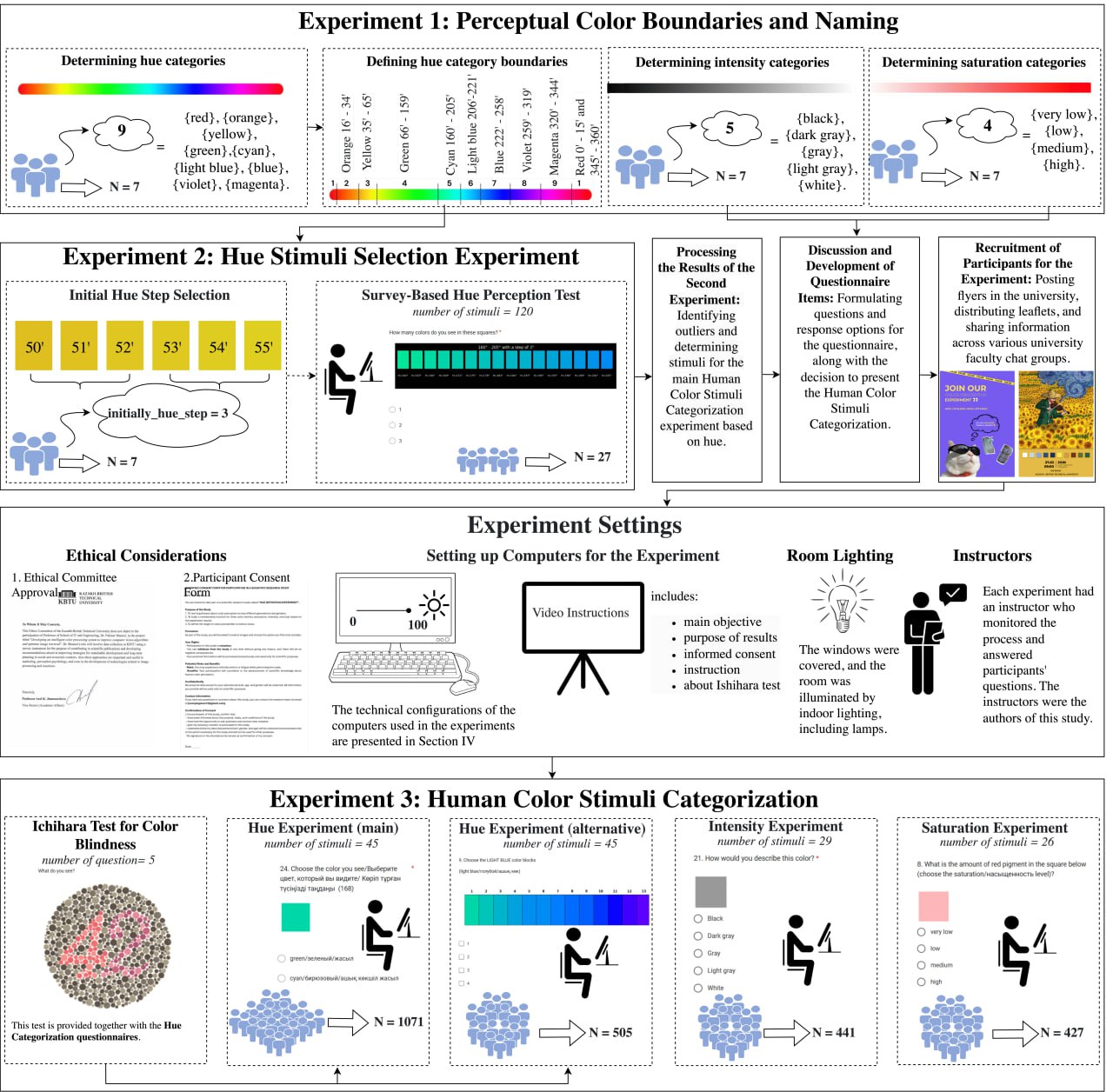}
  }
  \caption{Overview of the multistage experimental framework for constructing a human perception-based color model. The study consists of three main experiments: (1) Perceptual Color Boundaries and Naming, where participants defined hue, saturation, and intensity categories; (2) Hue Step Selection, which refined the stimuli for the final experiment; and (3) Human Color Stimuli Categorization, where participants classified color stimuli across different attributes. The experiment settings included ethical approvals, controlled lighting, and standardized computer configurations.}
  \label{fig:experiment}
\end{figure*}

A multistage experimental framework designed to construct a human perception-based color model is presented in Fig. \ref{fig:experiment}. It consists of three key experiments:
\begin{itemize}
    \item \textit{Perceptual Color Boundaries and Naming (Experiment 1).}Participants determined hue, saturation, and intensity categories, defining color boundaries and linguistic labels.
    \item \textit{Hue Stimuli Selection (Experiment 2)}. A survey was conducted to refine hue stimuli for the main experiment, ensuring an optimal selection for color categorization.
    \item \textit{Human Color Stimuli Categorization (Experiment 3)}. The main color categorization experiment involved participants classifying color stimuli based on hue, saturation, and intensity. The experiment also included an Ishihara test to assess color blindness.

\end{itemize}

An experimental setup was established in a Kazakh-British Technical University computer rooms to ensure controlled conditions. Each participant was seated comfortably at a standardized distance of 50–60 cm from the screen to maintain consistency in color perception (see Fig. \ref{fig:ins_1}). The participant pool consisted of bachelor’s, master’s, and PhD students, as well as academic faculty and staff.

Before the experiment began, each participant was clearly informed of the purpose, duration, and procedure. They were then given time to familiarize themselves with the interface and experimental setup before proceeding.

The experiment settings were carefully controlled, including lighting conditions, computer setup, ethical considerations, and instructor supervision, to maintain consistency across all trials (see Tab. \ref{tab:environment} and Fig. \ref{figure_SE}).

\subsubsection{Rooms Description}
\begin{figure*}[tb]
  \centering
  \makebox[\textwidth][c]{
 \includegraphics[width=0.9\textwidth]{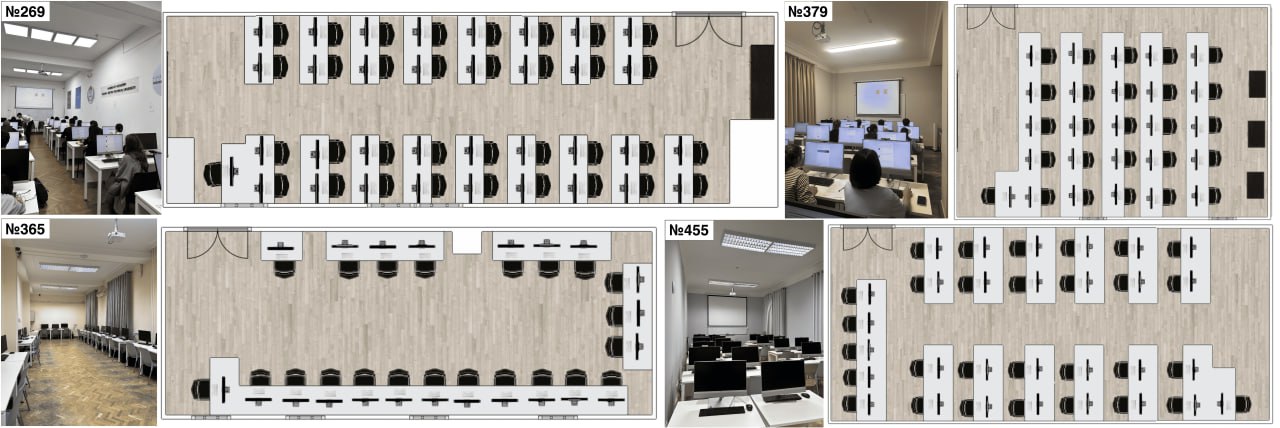}
  }
  \caption{
  Layouts and photographs of rooms 269, 265, 379, and 455 were used during the experiments. Diagrams illustrate the approximate arrangement of chairs, computers, windows, doors, and furniture. The number of workstations may slightly differ from the actual setup, as layouts reflect the configuration during the experiment and may have changed over time.}
  \label{fig:rooms_1}
\end{figure*}

The experiments were conducted in the university’s computer laboratories, numbered 269, 365, 379, and 455, on different days. During the experiments, all windows in the rooms were closed to maintain consistent lighting conditions. The rooms were relatively quiet, with no significant noise disturbances from the corridors. Each room contained two windows with curtains, ensuring the absence of reflective surfaces that could affect the results. The arrangement of computers, windows, and desks is presented in Fig.\ref{fig:rooms_1}. Key technical specifications are provided in Table \ref{tab:environment}. At the beginning of the experiment, participants took their assigned seats and then proceeded with the survey. The spacing between chairs allows students to move freely. This arrangement ensured that no participant obstructed another while moving to their respective places.  

Room 269 has white walls and 35 workstations, each with a computer, desk, and chair. The left-side windows are covered with curtains. The desk arrangement consists of two parallel rows, with two desks placed side by side in each row, leaving a central aisle for movement. 


In Room 365, the walls are painted in a light peach color. The room contains 25 workstations, each equipped with a computer and a desk. Upon entering, the projector and the instructor’s desk are immediately visible, followed by the participant workstations. The desks and computers are arranged along the walls, with their screens facing the center of the room and the backs of the computers against the walls. 

In Room 379, the walls are light gray. The room contains 25 workstations.  Upon entry, the projector, the instructor’s desk, and the first row of participant workstations are immediately visible. The desks are arranged in five rows, with five computers per row. There is ample space between the desks to allow movement, though the passage of individuals could be a potential distraction. To minimize disturbances, participants were assigned seats prioritizing unoccupied spaces. The windows are located parallel to the door on the right side of the instructor’s position.

In Room 455, the walls are light gray. The room is equipped with 25 workstations. The desks are arranged in five parallel rows, with two workstations on each side and an additional row positioned at the back of the room, featuring five workstations and computers. There is a wide passageway between the last and penultimate rows. The windows are arranged parallel to the door on the left side of the instructor’s desk. Due to the close arrangement of the two front rows, participants in the second row could be distracted by movement in the first row. To minimize disruptions, participants were instructed to occupy the first row before filling subsequent rows.

\begin{table*}[tb]
    \centering
    \renewcommand{\arraystretch}{1.2}
    \caption{Characteristics of classrooms and technical specifications of computers}
 \begin{tabular}{|l|l|p{1cm}|p{2cm}|p{2cm}|p{1cm}|p{3cm}|p{1.5cm}|}
        \hline
       
        \textbf{Room} & \textbf{Wall Color} & \textbf{Number of windows} & \textbf{Computer, OS} & \textbf{Software and Color Space} & \textbf{Area (m²)} & \textbf{Lighting Type} & \textbf{Number of workstations} \\
        \hline
        269 & White & 2 & ASUS,Windows & Windows Color Management and sRGB & 72,9 & Artificial, ceiling-mounted fluorescent lamps & 35 \\
        \hline
        365 & Light peach & 2 & ASUS,Windows & Windows Color Management and sRGB & 69,4 & Artificial, ceiling-mounted fluorescent lamps & 29 \\
        \hline
        379 & Light gray & 2 & HP, Windows & Windows Color Management and sRGB & 44,4 & Artificial, ceiling-mounted fluorescent lamps & 25 \\
        \hline
        455 & Light gray & 2 & HP, Windows & Windows Color Management and sRGB & 59,1 & Artificial, ceiling-mounted fluorescent lamps & 27 \\
        \hline
    \end{tabular}
    \label{tab:environment}
\end{table*}
\subsubsection{Multilingual Color Naming and Categorization}
\label{subsuc:Multilingual Color Naming and Categorization}



The study employs a multilingual approach, where all surveys and instructions were provided in Kazakh, Russian, and English (see Table \ref{tab:language}). 

Since participants had different mother tongues, their native language may have influenced their perception and categorization of colors. The use of Kazakh, English, and Russian in the experiment ensured that participants could interpret and rate colors in their most comfortable language, minimizing misunderstandings. The multilingual approach was chosen since, according to multiple studies, language shapes color perception, and categorization \cite{Josserand2021Environment, zgen2004Language, He2019Language, Siok2009Language}.

\begin{table*}[h!]
\centering
\renewcommand{\arraystretch}{1.2}
\caption{Translation of survey questions and color categories in English, Russian, and Kazakh. The multilingual design of the questions ensured consistent responses from participants, regardless of their native language}
\begin{tabular}{|c|c|c|c|}
\hline
\textbf{Language} &  \textbf{In English} &  \textbf{In Russian} &  \textbf{In Kazakh} \\
\hline
Question & Choose the color you see & \foreignlanguage{russian}{Выберите цвет, который вы видите} & \foreignlanguage{russian}{Көріп тұрған түсіңізді таңдаңыз} \\
\hline
Hue Categorization 1 & Red & \foreignlanguage{russian}{Красный} & \foreignlanguage{russian}{Қызыл} \\
\hline
Hue Categorization 2 & Orange & \foreignlanguage{russian}{Оранжевый} & \foreignlanguage{russian}{Қызғылт сары} \\
\hline
Hue Categorization 3 & Yellow & \foreignlanguage{russian}{Желтый} & \foreignlanguage{russian}{Сары} \\
\hline
Hue Categorization 4 & Green & \foreignlanguage{russian}{Зеленый} & \foreignlanguage{russian}{Жасыл} \\
\hline
Hue Categorization 5 & Cyan & \foreignlanguage{russian}{Бирюзовый} & \foreignlanguage{russian}{Ашық көкшіл жасыл} \\
\hline
Hue Categorization 6 & Light blue & \foreignlanguage{russian}{Голубой} & \foreignlanguage{russian}{Ашық көк} \\
\hline
Hue Categorization 7 & Blue & \foreignlanguage{russian}{Синий} & \foreignlanguage{russian}{Көк} \\
\hline
Hue Categorization 8 & Violet & \foreignlanguage{russian}{Фиолетовый} & \foreignlanguage{russian}{Күлгін} \\
\hline
Hue Categorization 9 & Magenta & \foreignlanguage{russian}{Розовый} & \foreignlanguage{russian}{Қызғылт} \\
\hline
\end{tabular}
\label{tab:language}
\end{table*}






\subsection{Experiment 1: Perceptual Color Boundaries and Naming }

\begin{figure}[tb]
 \includegraphics[width=0.45\textwidth]{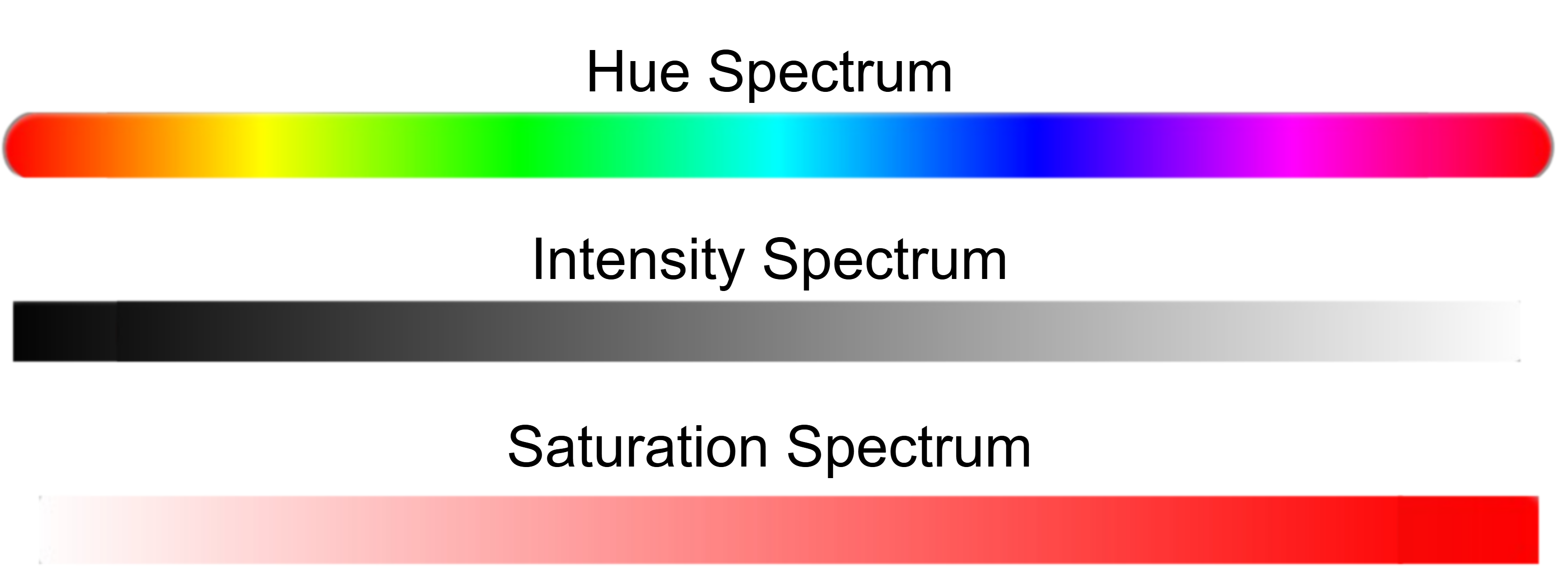}
 \caption{Spectrums for Hue, Intensity, and Saturation used in Experiment 1}
 \label{fig:spectrum}
\end{figure}

The goal of \textit{Experiment 1 } was to define linguistic color categories and perceptual boundaries for hue, intensity, and saturation using expert evaluation. A color spectrum for Hue, a grayscale spectrum for Intensity, and the red spectrum for Saturation are presented in Fig. \ref{fig:spectrum}.

\subsubsection{Hue Boundaries and Naming}

Seven color experts were presented with the full-color spectrum and asked to identify and categorize distinct hue groups based on their perceptions.
 The gradient ranged from red through the visible spectrum—progressing through orange, yellow, green, cyan, blue, and violet—before looping back to red. Each participant individually observed the gradient and marked where they believed the boundaries of distinct colors lay. They also assigned names to the colors they perceived within those boundaries.



We averaged the participants' perceptions of color segmentation and received nine linguistic hue categories: \textit{Red, Orange, Yellow, Green, Cyan, Light Blue, Blue, Violet}, and \textit{Magenta}. 

Based on these experimental results, \textit{Light blue} appeared as a distinct color category.  In many languages, the term for\textit{ Light blue }is treated as a separate and distinct color category from blue. In contrast, \textit{Light blue} is often regarded as a variation of \textit{Blue }rather than a separate category in other languages. By including \textit{Light blue }as a standalone category, we aim to account for these cultural nuances to get a more accurate representation of color perception across diverse populations. This is also supported by cross-linguistic and cultural differences in color perception and naming \cite{Jonauskaite2020feeling}.


After establishing linguistic categories, the precise boundaries for each hue were determined using results from \cite{fuzzycolor_artemotion}. These boundaries define the initial range of hues corresponding to each category, which is necessary for \textit{Experiment 2}.

\subsubsection{Intensity Boundaries and Naming}
\begin{itemize}
    \item A grayscale spectrum (ranging from black to white) was shown to seven experts to define intensity (brightness) levels.
    \item A chromatic-neutral approach was used to avoid color bias.
    \item Five intensity categories were established: \textit{Black, Dark gray, Gray, Light gray, White}
\end{itemize}

\subsubsection{Saturation Boundaries and Naming}
\begin{itemize}
    \item Since saturation cannot be assessed without color, a red spectrum (ranging from desaturated white to fully saturated red) was displayed to seven experts.
    \item The experts categorized saturation into four levels: \textit{Very low, Low, Medium, High}
    \item Red was chosen as the reference color due to its popularity and the fact that the Hue range starts from this color.
\end{itemize}

This experiment provided the fundamental color categories necessary for further research, forming the basis for the selection of hue stimuli and human color categorization experiments.

\subsection{Experiment 2: Hue Stimuli Selection Experiment}

\begin{figure*}[th]
  \includegraphics[width=\textwidth]{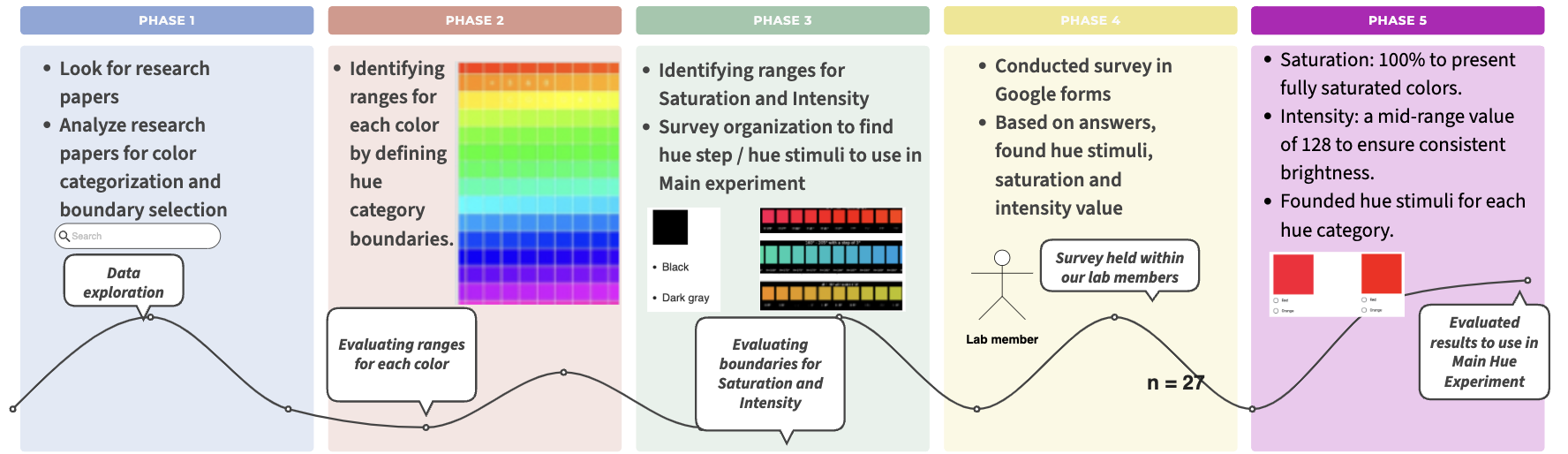}
  \caption{Overview of Experiment 2: selecting hue stimuli with specific saturation and intensity values for the main experiment.}
  \label{fig:exp2}
\end{figure*}
This is a preliminary experiment that aims to establish the foundational color stimuli for the main color categorization experiment. The overview of\textit{ Experiment 2 }is provided in Fig. \ref{fig:exp2}. We chose fixed values for saturation and intensity: the maximum value for saturation (S=75\%) and the average value for intensity (I=50\%). 

 \subsubsection{Initial Hue Step Selection}
    \begin{itemize}
        \item The 360-degree hue spectrum was initially divided into hue steps of 3 degrees, resulting in 120 color stimuli.
        \item A small group of authors and experts evaluated this initial selection, ensuring that the stimuli covered a balanced range of hues within the predefined linguistic labels from \textit{Experiment 1} and boundaries from \cite{fuzzycolor_artemotion}.
    \end{itemize}

  \subsubsection{Survey-Based Hue Perception Test}
    A group of 27 people participated to ensure an initial understanding of human responses to various color stimuli. Those are master's, bachelor's, and PhD students from Kazakh-British Technical University. 
    \begin{itemize}
    \item The experiment was conducted as a survey. The survey included nine questions, each corresponding to one of the nine color categories. 
        \item Participants were shown a series of color stimuli containing hues sampled at three-degree steps (total 120 hues) within the predefined nine hue category boundaries. Each sequence was taken from the range of one of the nine color categories. 
        \item For each range, they were asked: \textit{"How many distinct colors do you see in this set?"} to determine the number of perceptually distinguishable colors in each boundary range. For example, if the range pertains to the red category, participants note how many hues they can distinguish within that range. 
        \item The results were processed, and the stimuli for the main Human Color Stimuli Categorization Experiment were finalized by dividing the range according to the number of visible colors.
    \end{itemize}
The range for each color category was defined based on \cite{fuzzycolor_artemotion}. For red, it’s from 0 to 15 and 345 to 360; for orange, 16 to 34; for yellow, 35 to 65; for green, 66 to 159; for cyan, 160 to 205; for light blue, 206 to 221; for blue, 222 to 258; for violet, 259 to 319; and for magenta, 320 to 344.

As a result, we can identify the steps and the stimuli to be used in the \textit{Experiment 3}. To calculate the step, we divided the range by the average value obtained from participants' responses (see Eq. \ref{eq:step_size}). This process was repeated for each of the nine color categories. The resulting value was referred to as the step. For example, the step number for the red category was seven, so we selected every 7$^{th}$ hue from the red range in \textit{the Experiment 3}. 

\begin{equation}
\text{Step size} = \frac{H_{\text{range}}}{\bar{N}_{\text{hues}}}
\label{eq:step_size}
\end{equation}

where:
\begin{itemize}
    \item \( H_{\text{range}} \) is the total hue range for the given color category.
    \item \( \bar{N}_{\text{hues}} \) is the average number of distinguishable hues as perceived by participants.
\end{itemize}

The results from this preliminary experiment are distinguishable color stimuli for hue. These stimuli will serve as the reference points for the main experiment.  This experiment was critical in defining the final color stimuli set, ensuring that hue steps were perceptually meaningful. The intensity and saturation stimuli were selected uniformly, resulting in 29 and 26 stimuli, respectively.

\subsection{Experiment 3 (Main): Human Color Stimuli Categorization}
In the previous step, we identified key hue angles that participants perceived as distinct enough for categorization. This is the main experiment of the study: human color categorization through direct rating and alternative settings. Building on the results of previous experiments, the main experiment investigates how humans categorize colors in terms of hue, saturation, and intensity.

The experiment was conducted using the Hue-Saturation-Intensity (HSI) color space with the following parameters:
\begin{itemize}
    \item \textit{Saturation.} Maintained at 100\% to present fully saturated colors.
    \item \textit{Intensity.} Fixed at a mid-range value of 128 to ensure consistent brightness.
    \item \textit{Hue.} The only variable parameter, covering the complete 360$^{\circ}$ spectrum.
\end{itemize}

The Hue categorization experiment was designed in two versions to support both forced-choice categorization and flexible selection approaches.:
        \begin{itemize}
            \item \textit{Main Version} – required participants to assign a single category to each stimulus.
            \item \textit{Alternative Version} – allowed participants to select multiple categories for each stimulus or leave stimuli unclassified, providing more flexibility in color labeling.
        \end{itemize}

The call for participation was disseminated through flyers, faculty chat groups, emails, and social media, inviting a diverse pool of university participants to contribute to the experiment.
   
The experiment begins with instructions and an explanation of the questions' structure and the response format. Participants sat at a computer and watched a video instruction on an interactAdditionally, the video emphasized the importance of reviewing and signing the informed consent form esults, and detailed instructions, including maintaining eye level with the screen, the required distance from the monitor, the importance of not looking at others' responses, responding quickly, and using "\_" or "nothing" if nothing is visible during an Ishihara test for color blindness. 

In addition, the video emphasized the necessity of reviewing the informed consent form and signing it before proceeding. On the tables next to each computer, printed copies of the informed consent form were available in three languages (English, Kazakh, and Russian). All participants provided informed consent, and their anonymity was maintained.  If it is a Hue Categorization experiment, instructions are followed by the Ishihara test for color blindness.

The following section is a questionnaire that includes questions about the participant's role. The participant may be a bachelor's student, a master's student, a doctoral student, or a faculty member (e.g., a professor). This classification was designed with the understanding that the experiment is conducted in a university setting. Other questions were related to the participants' age and gender to analyze the impact of demographic factors on perception.


Following the questionnaire, both Hue Categorization experiments present the same set of 45 (1, 5, 12, 15, 20, 25, 30, 34, 39, 44, 49, 54, 59, 65, 81, 97, 113, 129, 145, 159, 168, 177, 186, 195, 204, 205, 211, 217, 221, 231, 241, 251, 258, 270, 282, 294, 306, 318, 319, 327, 335, 343, 344, 351, 358) color stimuli in different formats. In contrast, saturation categorization includes a different set of 26 stimuli, while intensity categorization features another distinct set of 29 stimuli. A detailed description of the stimuli and their presentation methodology is provided in Table \ref{tab:exp_conditions}.

All participants voluntarily participated in the survey and provided demographic information, including their age, gender, and educational level. Before participating, all agreed to provide informed consent and completed a color blindness test to ensure accurate color perception. After that, they read the survey instructions and then started the main survey on color perception. Settings such as lighting, screen calibration, and viewing distance were standardized across all participants(see Fig. \ref{fig:ins_1} and Tab.\ref{tab:environment}).

\subsubsection{Hue Categorization (main settings)}

This experiment aimed to collect data needed to build fuzzy membership functions that represent human perception of hue, saturation, and intensity. 



\begin{figure*}[tb]
\centering
\begin{subfigure}[h]{0.6\textwidth}
    \centering
    \includegraphics[width=\textwidth]{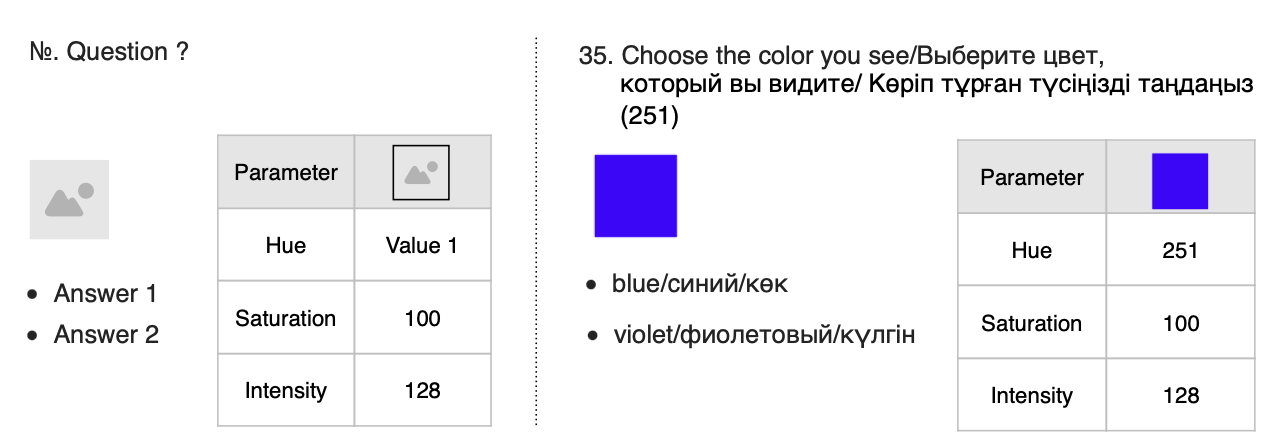} 
    \caption{Hue (main) Survey Design and Example with Color Stimuli}
    \label{fig:img-a}
\end{subfigure}
\hfill
\begin{subfigure}[h]{0.36\textwidth}
    \centering
    \includegraphics[width=\textwidth]{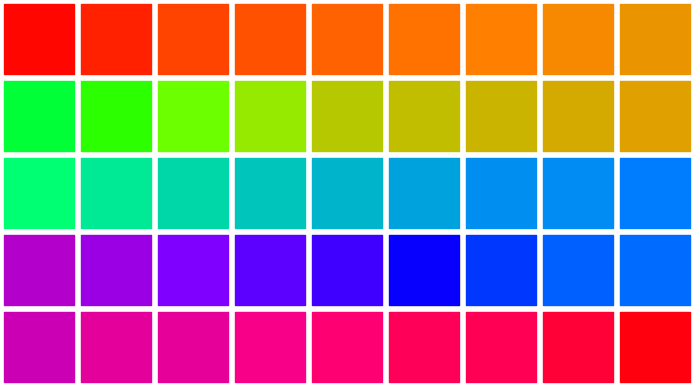 } 
    \caption{All color stimuli}
    \label{fig:img-b}
\end{subfigure}
\caption{The design of the question formulation for Experiment 3: Hue Categorization with an example of a single color stimulus and all the stimuli that were used in the experiment}
\label{fig:mockup_all_stimuli}
\end{figure*}



  
  
  
  

\paragraph{Participants} 

The total number of participants was 1,071, including 997 bachelor's students (93.1\%), 29 master's students (2.7\%), 11 doctoral students (1\%), and 34 professors (3.2\%). The gender distribution was 40.3\% female (432 participants) and 59.7\% male (639 participants). 

The age distribution of participants was as follows: Under 18 years - 61 participants (5.7\%), 18-24 years - 960 participants (89.6\%), 25-34 years - 26 participants (2.4\%), 35-44 years - 15 participants (1.4\%), and 45-54 years - 8 participants (0.7\%).

 
  
\paragraph{Experimental Design}

This experiment consists of five Ishihara test questions, followed by three questions regarding participant information, and 45 hue categorization questions. These questions present color stimuli as square PNG images with a resolution of 77×77 pixels. Fig.\ref{fig:mockup_all_stimuli} illustrates the structure of the questionnaire and presents all the stimuli included in the survey.

The survey was open on the computers, and the questionnaire was divided into sections. The first section provides instructions on the Ishihara test, and the following five sections contain the Ishihara test. The sixth section includes a brief description of the survey and an expression of gratitude for participation. After this, the main 48 questions are presented.

All questions and response options are available in three languages: English, Russian, and Kazakh (the reasons for this are detailed in Section \ref{subsuc:Multilingual Color Naming and Categorization} and Table\ref{tab:language}). Each question is numbered, displaying one image and two answer choices.

The details regarding the room setup and computer configurations are outlined in Section \ref{subsec:Experiment Settings}. 

\paragraph{Experimental Procedure}




\begin{itemize}
\item \textit{Demographic Information Collection}. Participants selected their age, gender, and educational background from predefined options in the survey, allowing for analysis of potential differences in color perception.
\item \textit{Ishihara test for color blindness}.
\item \textit{Color Perception Task}.Participants were shown isolated color stimuli and asked to choose between two options representing different hues.
\item \textit{Response Analysis}.The recorded responses were analyzed to understand how individuals perceive color hues and to contribute to the future development of a color categorization model.
\end{itemize}

The results of this experiment will contribute to the construction of a membership function, which will enable the assessment of the degree to which a given color belongs to a specific category.




  

 
 

\subsubsection{Hue Categorization (alternative settings)}

This experiment investigates how humans perceive color boundaries within continuous gradients. The research specifically focuses on identifying where people subjectively distinguish between different colors when only the hue varies while saturation and intensity remain constant. It can help to clarify how individuals categorize colors and where they perceive transitions between different color categories.

\begin{figure*}[tb]
\centering
\begin{subfigure}[h]{0.49\textwidth}
    \centering
    \includegraphics[width=\textwidth]{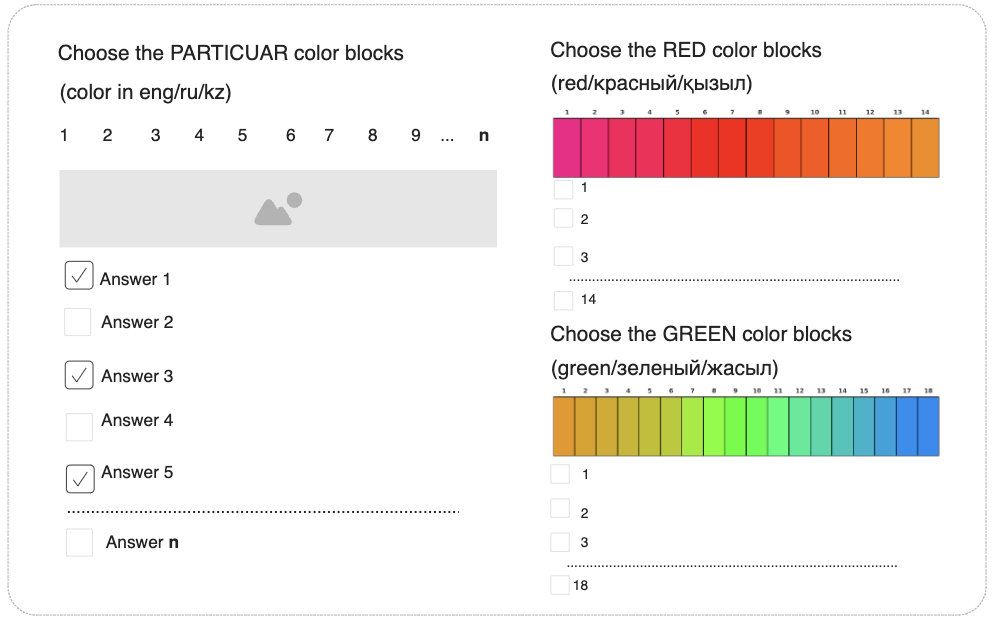} 
    \caption{Hue (alternative) Survey Design and Example with Colors.}
    \label{fig:imgalt-a}
\end{subfigure}
\hfill
\begin{subfigure}[h]{0.49\textwidth}
    \centering
    \includegraphics[width=\textwidth]{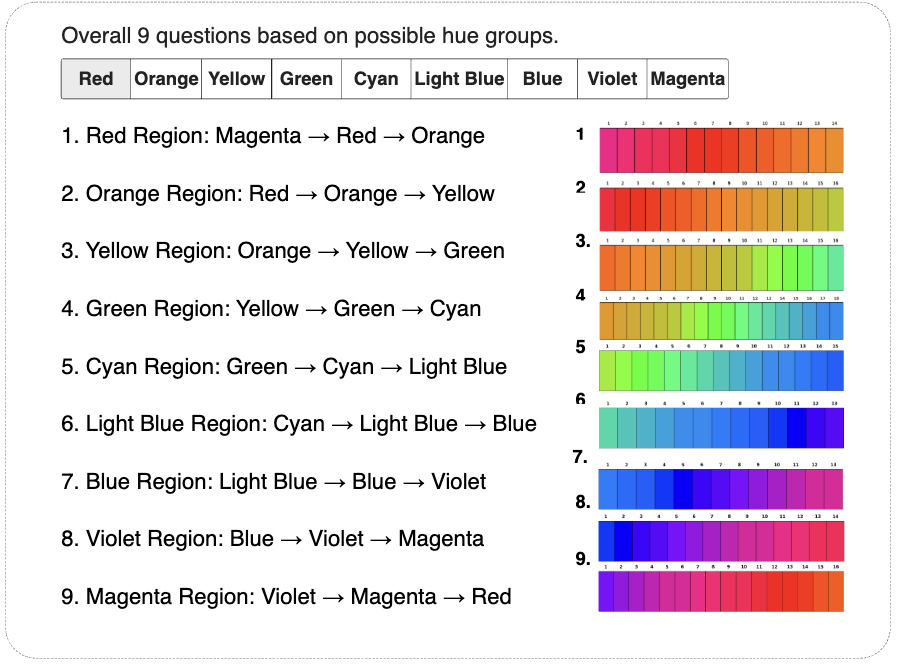} 
    \caption{All hue transitions illustrated.}
    \label{fig:imgalt-b}
\end{subfigure}
\caption{The interface of Hue (alternative) Experiment. Participants identify color boundaries in a structured selection task.}
\label{fig_alternative_hue}
\end{figure*}

We conducted two types of surveys for our main \textit{Hue} experiment. This Hue experiment (alternative) consisted of blocks of colors, where the user should choose the specific blocks of a particular color (see Fig. \ref{fig_alternative_hue}). This type of question explores how the user perceives colors in a seamless gradient.

The experiment aimed to investigate human perception of colors within a gradient, focusing on participants' ability to identify transitions between specific colors (e.g., red, blue, etc.). This survey included five Ishihara test questions followed by nine questions about possible hue gradients.

\paragraph{Participants}

A total of 505 individuals participated in this study. The sample consisted of 267 males (52.87\%) and 238 females (47.13\%), ensuring a balanced gender distribution.
In terms of academic roles, the majority of participants were bachelor's students (\textit{n} = 441, 87.33\%), followed by professors (\textit{n} = 30, 5.94\%), master's students (\textit{n} = 25, 4.95\%), and PhD students (\textit{n} = 9, 1.78\%).
The age distribution of the participants was as follows: 18-24 years (\textit{n} = 436, 86.34\%), 25-34 years (\textit{n} = 28, 5.54\%), under 18 years (\textit{n} = 20, 3.96\%), 35-44 years (\textit{n} = 12, 2.38\%), 45-54 years (\textit{n} = 8, 1.58\%), and 55-64 years (\textit{n} = 1, 0.20\%). So, we have a diverse participant pool, primarily composed of young adults in academia.



\paragraph{Experiment Design}
Fig. \ref{fig_alternative_hue} illustrates the setup of the experiment. Fig. \ref{fig:imgalt-a} presents the survey mock-up, showing how participants categorized a single color stimulus, while Fig. \ref{fig:imgalt-b} displays the complete set of color stimuli used in the experiment.

The participants saw a row of color blocks, each numbered from 1 to 14/16/18, above the block. These blocks represented a smooth color gradient that transitioned from one color to another. For instance, we need to consider the red color's left and right neighbors, so the stimuli rectangle starts with magenta colors, then red, then orange. The hue groups were structured to examine all possible transitions.

The survey interface featured smooth hue transitions with numbered blocks, enabling participants to define their subjective color boundaries. The experiment mock-up and the settings can be seen in Figure \ref{fig_alternative_hue}.

By carefully examining the blocks, each person selects all the blocks that, in their opinion, appear to be red or a specific color. In addition, an individual can choose one block, multiple blocks, or no blocks if they feel none represents a particular color. 

The experimental stimuli consisted of the same set of colors, each representing a different hue value (see Fig. \ref{fig:img-b}). Each block was assigned a unique identification number. This numbering system facilitated:

\begin{enumerate}
    \item Clear identification of each distinct hue value.
    \item Systematic selection using sequential IDs displayed above each block.
    \item Straightforward recording of participant selections as numerical ranges.
\end{enumerate}

\paragraph{Experimental Procedure}
The experiment followed these steps:

\begin{itemize}
\item \textit{Demographic Information Collection}. Participants provided data on their age, gender, and educational background to analyze potential variations in color perception.
\item \textit{Ishihara test for color blindness}.
    \item \textit{Color Perception Task}. Participants were shown a series of colored blocks and asked to choose which blocks seemed a certain color to them, depending on the color that was asked.

\item \textit{Response Analysis}. Responses were recorded as selections of numbered blocks, indicating which blocks participants associated with each hue category.

\end{itemize}




\subsubsection{Saturation Categorization}

During the next experiment, participants completed a survey by identifying the saturation level of a given color stimulus. This study examines how people perceive saturation when the hue is set to zero and the intensity is at a mid-level. Understanding saturation perception helps us learn more about human vision and color recognition.
\paragraph{Participants}
A total of 427 participants voluntarily took part in the survey, comprising 214 females (50.11\%) and 213 males (49.89\%). Those are master's (\textit{n} = 127, 29.74\%), bachelor's (\textit{n} = 270, 63.23\%), PhD students (\textit{n} = 6, 1.41\%), and faculty (n = 24, 5.62\%) from Kazakh-British Technical University.  
\paragraph{Experiment Design}
\begin{figure*}[th]
    \centering
    \includegraphics[width=0.9\linewidth]{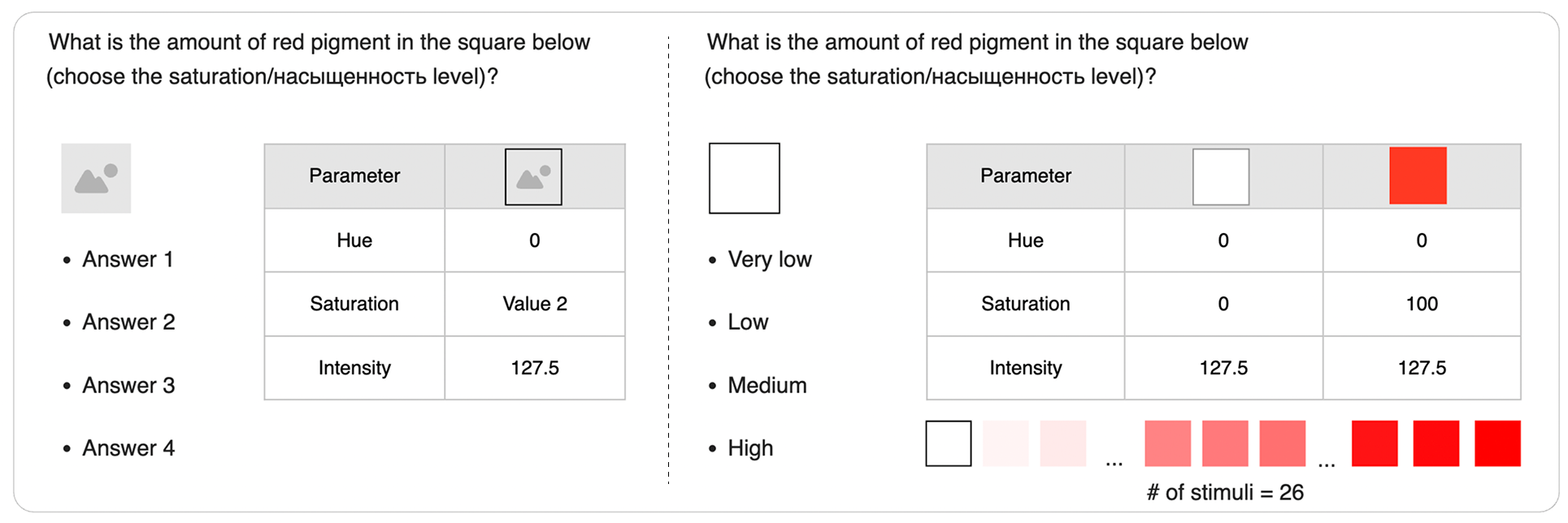}
    \caption{Saturation survey design and example with color stimuli}
    \label{fig:sat_mock-up}
\end{figure*}
The experimental design for Saturation categorization is presented in Fig. \ref{fig:sat_mock-up}. The participants were required to answer 26 questions, categorizing each color stimulus into one of four groups according to saturation: \textit{Very low, Low, Medium}, or \textit{High}. Specific parameters were defined in the Hue-Saturation-Intensity (HSI) color space: 
\begin{itemize}
    \item \textit{Hue}. The hue value is set to zero
    \item \textit{Saturation.} The saturation value varies with a step of four.
    \item \textit{Intensity.} The intensity value is fixed at 50\%, enabling mid-intensity.
\end{itemize}
We chose these specific values to observe how the color changes according to saturation values. The zero hue ensures that color remains consistent throughout the experiment, as the hue determines the specific position in the color spectrum. We eliminated variations in the base color. The middle-intensity value provides a neutral base, avoiding extremes of color brightness that could distract participants from perceiving saturation changes. The four-step incremental change provides a smooth and consistent progression from near-white (low saturation) to vivid red (high saturation), as shown in Fig. \ref{fig:sat_mock-up}.

\paragraph{Experimental Procedure}
The saturation categorization experiment included the following steps:
\begin{itemize}
    \item \textit{Demographic Information Collection.} Participants provided information about gender, age, and education for further analysis.
    \item \textit{Saturation Perception Task.} Participants identified the saturation level according to their perception and classified the color stimuli into categories. 
    \item \textit{Response Analysis.} The recorded responses were analyzed to understand how individuals perceive saturation and to contribute to the future development of a color model. 
\end{itemize}


\subsubsection{Intensity Categorization}
\begin{figure*}[th]
    \centering
    \includegraphics[width=0.9\linewidth]{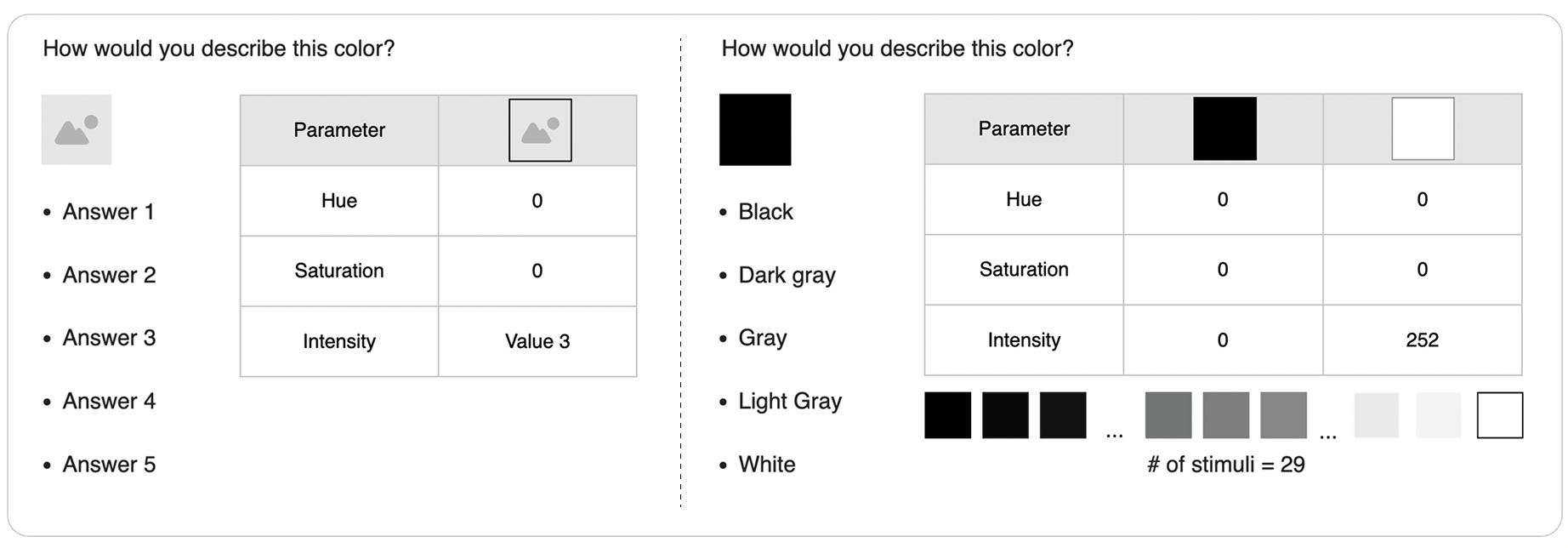}
    \caption{Intensity survey design and example with color stimuli}
    \label{fig:int_mock-up}
\end{figure*}

Intensity Categorization Experiment Design is shown in Fig. \ref{fig:int_mock-up}. In this experiment, participants categorized color stimuli based on intensity levels. Therefore, this study will help explore how people perceive and distinguish intensity.
\paragraph{Participants}
A total of 441 individuals participated in the experiment, comprising 220 females (49.89\%) and 221 males (50.11\%). Those are master's (\textit{n} = 128, 29.02\%), bachelor's (\textit{n} = 282, 63.95\%), PhD students (\textit{n} = 6, 1.36\%), and faculty (\textit{n} = 25, 5.67\%) from Kazakh-British Technical University.  

\paragraph{Experiment Design}
The experiment survey included 29 questions. Participants were asked to categorize the stimuli into the following categories: Black, Dark Gray, Gray, Light Gray, and White.
\begin{itemize}
    \item \textit{Hue.} The hue value is set to zero.
    \item \textit{Saturation.} The saturation is fixed at zero.
    \item \textit{Intensity.} The intensity value increases by nine.
\end{itemize}
The experimental settings were designed to create a clear range of grayscale shades, making it easier for participants to categorize them. We removed color by setting hue and saturation to zero, so the focus is only on the intensity. A step of nine was chosen to ensure sufficient variation between shades while maintaining consistent differences. This setup enables us to test how well participants can distinguish and categorize different levels of color intensity.

\paragraph{Experimental Procedure}
The intensity categorization experiment included the following steps:
\begin{itemize}
    \item \textit{Demographic Information Collection.} Participants provided information about gender, age, and education for further analysis.
    \item \textit{Intensity Perception Task.} Participants identified the intensity level according to their perception and classified the achromatic color stimuli into categories. 
    \item \textit{Response Analysis.} The recorded responses were analyzed to understand how individuals perceive intensity and to contribute to the future development of a color model. 
\end{itemize}

\section{Results and Analysis}

\label{sec:modelling}
\subsection{Experimental Results and Analysis} 

\subsubsection{Hue Categorization Analysis}


\begin{figure}[ht!]
    \centering
    \includegraphics[width=\linewidth]{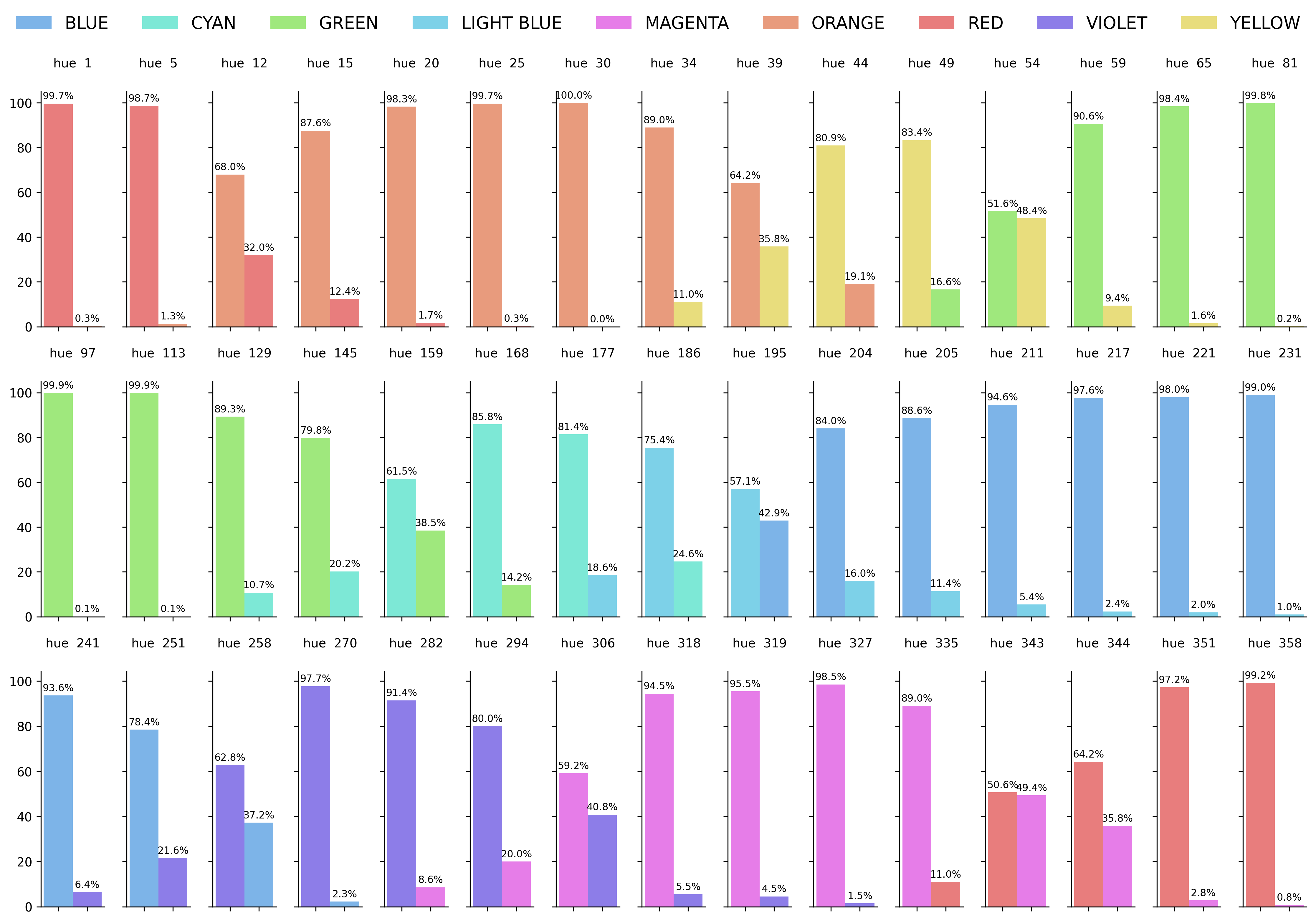}
    \caption{The distribution of (main) hue perception responses for each color stimulus.}
    \label{main-bar}
\end{figure}

The Fig. \ref{main-bar} illustrates the results of the main hue experiment after removing outliers, including color blind respondents, in which participants were asked to classify 45 stimuli into the given two color categories. Each bar graph shows the percentage distribution of responses for each stimulus.


According to the results, participants generally demonstrated a strong consensus on identifying the stimuli. Several stimuli, such as 1, 5, 20, 25, 30, 65, 81, 97, 113, 217, 221, 231, 270, 319, 327, 351, and 358 show a dominant response exceeding 95\%, indicating an unambiguous perception. These stimuli are likely to be highly identifiable, indicating minimal variation in subjective perception.


Some bar charts indicate small but noticeable misclassifications, with 5-10\% of responses deviating significantly from the expected color category (e.g., stimuli 59, 211, 241, 282, 318). While these cases are rare, they highlight individual perceptual differences or potential human error in the selection process.


Certain stimuli demonstrate moderate disagreement, where the dominant choice is evident, yet alternative responses appear more frequently. For instance, in column 251, the primary color is identified by 78.4\% of participants, while 21.6\% selected a different classification. Similar patterns emerge in stimuli 44, 49, 145, 177, 186, and others, where alternative color choices account for up to a third of responses. These instances suggest either color similarity between the given option and another choice or individual differences in color perception.


A few stimuli received nearly equal proportions of responses across two categories, as seen in stimuli 54 and 343, where the responses were divided  51.6\% to 48.4\% and  50.6\% to 49.4\%, respectively. Such cases indicate uncertain color perception, where stimuli may lie on the boundary between two categories. 

Overall, the results demonstrate that most colors are perceived consistently, with clear majorities identifying the color category. However, a subset of colors exhibits moderate to substantial ambiguity, suggesting potential challenges in consistent color naming.

\begin{figure}[tb]
    \centering
    \includegraphics[width=\linewidth]{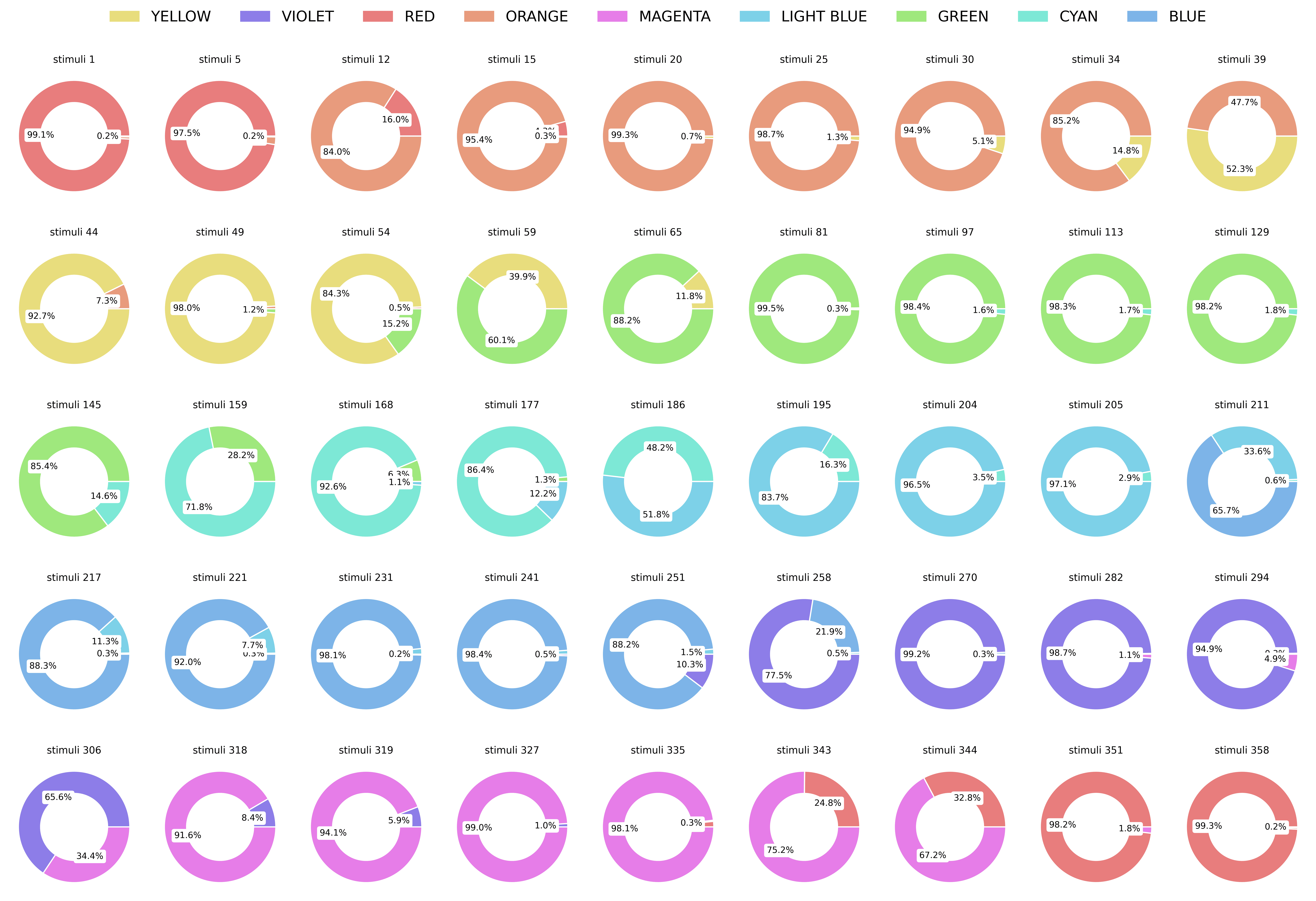}
    \caption{The distribution of (alternative) hue perception responses for each color stimulus.}
    \label{alt-pie}
\end{figure}

 In the Hue alternative categorization experiment, respondents were asked to select which stimuli belonged to each color category. We then calculated the number of stimuli chosen per color group for each respondent and compared the results by gender. The analysis showed minimal differences between the male and female participants, as shown in Table \ref{tab:hue_alter_stimule_count}. Mean values for most color categories varied by only 0.1 to 0.53, indicating similar categorization behavior. For example, males selected slightly more stimuli in the red category (Male: 3.74; Female: 3.30), while females categorized slightly more stimuli in the blue category (Female: 4.01; Male: 3.83). The green category had the highest average stimulus distribution in both genders (Female: 5.21; Male: 4.68). The data from the experiment do not indicate any significant gender-specific trends in the distribution of stimuli for color categorization. Overall, the similarity in stimulus distribution between male and female respondents suggests a shared conceptual distribution of color categories, at least within the study's experimental context and cultural framework.

\begin{table}[]
\caption{Distribution of Color Stimuli Categorization by Gender (Hue Alternative experiment)}
\label{tab:hue_alter_stimule_count}
\begin{tabular}{|c|rrr|rrr|}
\hline
\multicolumn{1}{|l|}{} & \multicolumn{3}{c|}{\textbf{Female - 233}}              & \multicolumn{3}{c|}{\textbf{Male - 254}}                \\ \hline
\multicolumn{1}{|l|}{} &
  \multicolumn{1}{c|}{\textbf{max}} &
  \multicolumn{1}{c|}{\textbf{mean}} &
  \multicolumn{1}{c|}{\textbf{min}} &
  \multicolumn{1}{c|}{\textbf{max}} &
  \multicolumn{1}{c|}{\textbf{mean}} &
  \multicolumn{1}{c|}{\textbf{min}} \\ \hline
\textbf{blue}          & \multicolumn{1}{r|}{7}  & \multicolumn{1}{r|}{4.01} & 1 & \multicolumn{1}{r|}{8}  & \multicolumn{1}{r|}{3.83} & 1 \\ \hline
\textbf{cyan}          & \multicolumn{1}{r|}{5}  & \multicolumn{1}{r|}{2.58} & 0 & \multicolumn{1}{r|}{6}  & \multicolumn{1}{r|}{2.92} & 1 \\ \hline
\textbf{green}         & \multicolumn{1}{r|}{10} & \multicolumn{1}{r|}{5.21} & 1 & \multicolumn{1}{r|}{10} & \multicolumn{1}{r|}{4.68} & 1 \\ \hline
\textbf{light blue}    & \multicolumn{1}{r|}{9}  & \multicolumn{1}{r|}{2.45} & 0 & \multicolumn{1}{r|}{9}  & \multicolumn{1}{r|}{2.76} & 0 \\ \hline
\textbf{magenta}       & \multicolumn{1}{r|}{8}  & \multicolumn{1}{r|}{3.70} & 0 & \multicolumn{1}{r|}{7}  & \multicolumn{1}{r|}{3.23} & 1 \\ \hline
\textbf{orange}        & \multicolumn{1}{r|}{10} & \multicolumn{1}{r|}{4.75} & 1 & \multicolumn{1}{r|}{8}  & \multicolumn{1}{r|}{4.44} & 1 \\ \hline
\textbf{red}           & \multicolumn{1}{r|}{7}  & \multicolumn{1}{r|}{3.30} & 1 & \multicolumn{1}{r|}{7}  & \multicolumn{1}{r|}{3.74} & 1 \\ \hline
\textbf{violet}        & \multicolumn{1}{r|}{6}  & \multicolumn{1}{r|}{3.09} & 0 & \multicolumn{1}{r|}{7}  & \multicolumn{1}{r|}{3.25} & 1 \\ \hline
\textbf{yellow}        & \multicolumn{1}{r|}{6}  & \multicolumn{1}{r|}{1.92} & 0 & \multicolumn{1}{r|}{7}  & \multicolumn{1}{r|}{2.07} & 0 \\ \hline
\end{tabular}
\end{table}

\subsubsection{Saturation Categorization Analysis}

\begin{figure*}[tb]
\centering
\begin{subfigure}[h]{0.49\textwidth}
    \centering
    \includegraphics[width=\textwidth]{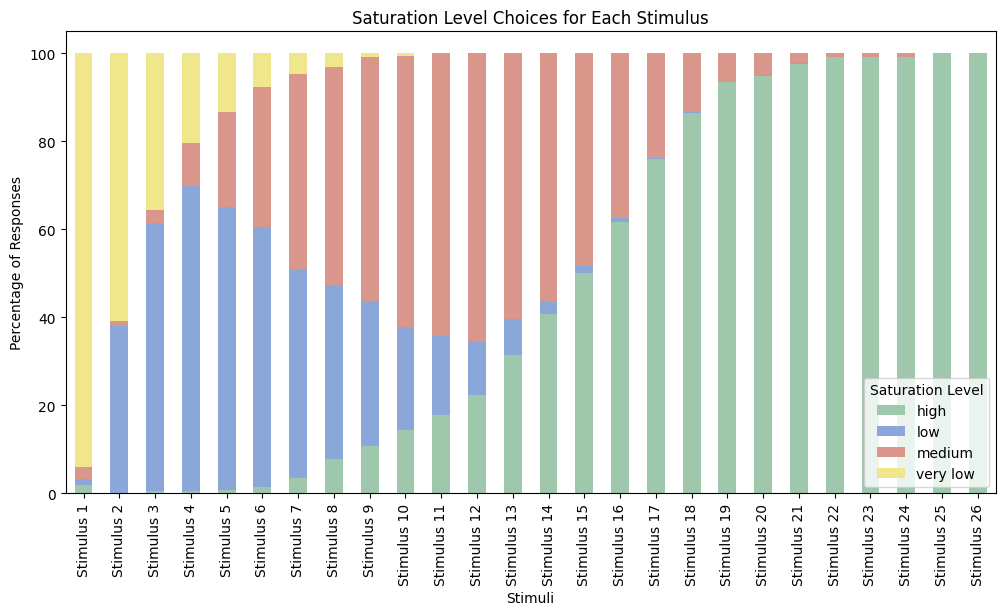} 
    \caption{Saturation Survey Results}
    \label{fig:img-sat}
\end{subfigure}
\hfill
\begin{subfigure}[h]{0.49\textwidth}
    \centering
    \includegraphics[width=\textwidth]{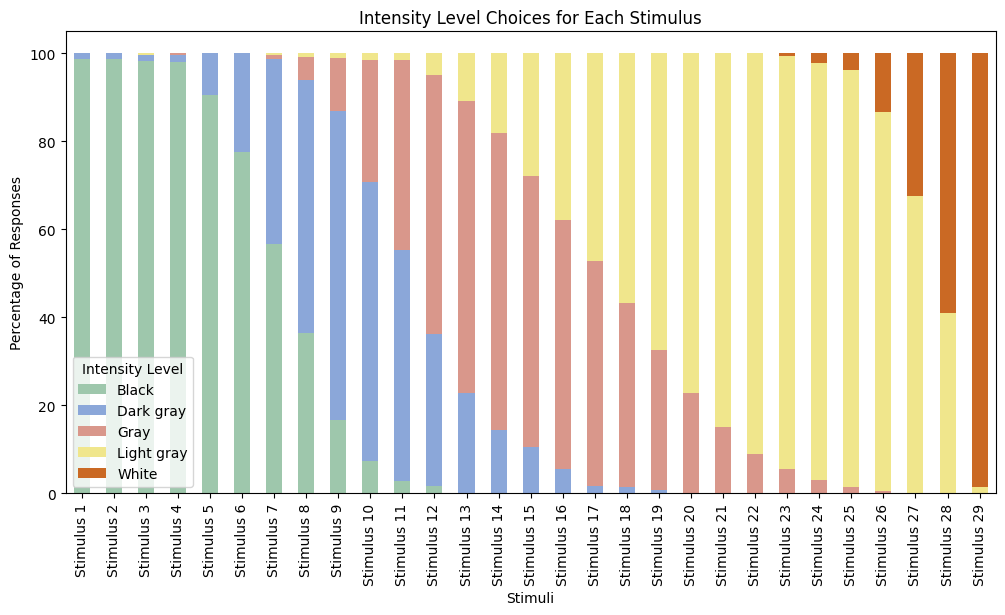} 
    \caption{Intensity Survey Results}
    \label{fig:img-int}
\end{subfigure}
\caption{The distribution of saturation and intensity perception responses}
\label{fig:sat_int_results}
\end{figure*}

The Fig. \ref{fig:img-sat} presents the results of the survey on saturation perception. The stacked bar chart illustrates how the participants classified each of the 26 stimuli. This diagram illustrates the distribution of participants' responses, showing agreement levels and potential ambiguity across saturation levels.

As can be seen, participants largely shared the same opinion for stimuli 1 - 5 and stimuli 16 - 26. This suggests that these stimuli were more distinct and aligned with typical human perception. However, Stimuli 7 and 8 caused ambiguity among participants in choosing between the "low" and "medium" categories. This could be due to difficulty distinguishing between mid-saturation levels, as extreme categories are easier to judge, or to external factors such as lighting, screen settings, and individual perception.

In general, despite some stimuli appearing ambiguous, participants consistently perceived and classified saturation levels uniformly. The survey observations provide insights into how people perceive saturation levels and identify areas of uncertainty. This can help optimize color usage in design, visual displays, and user interfaces for better clarity and differentiation.

\subsubsection{Intensity Categorization Analysis}

The Fig.\ref{fig:img-int} represents the distribution of responses in the intensity survey. Each bar is composed of multiple segments, with each segment representing a distinct level of response intensity. Color-coded segments allow for easy comparison of the intensity distribution between categories.

According to the diagram, most participants responded similarly at the beginning and end of the survey, primarily classifying stimuli based on monochrome intensity levels. Their responses were mainly limited to black and white.

However, for stimuli 7 and 11, there is noticeable ambiguity in the decision between black and dark gray, as well as between dark gray and gray. Various factors, such as individual perception of intensity, lighting conditions, or differences in computer display settings, may cause this uncertainty.

Overall, participants generally agreed on stimulus intensity, except for a few cases. 


\subsubsection{Color Blind Test Analysis}

Our model is particularly inclusive for color-blind individuals, as it provides linguistic descriptions of colors rather than numerical values. Unlike traditional models, which assume uniform color perception, our fuzzy-based approach adapts to varying perceptions of color. It allows users with color vision deficiencies to interpret and categorize colors through descriptive labels, making it the only model that effectively supports diverse visual experiences. 

Fig.  \ref{fig:preprocessing} presents results from the \textit{Hue Main }and \textit{Hue Alternative} experiments of the Ishihara Test. In the \textit{Hue Main} experiment, a total of 1,071 participants took the test. Among them, 22 respondents left at least one question unanswered, while 33 provided at least one incorrect response. Among these participants, 76.4\% were male and 23.6\% were female. Similarly, in the \textit{Hue Alternative }experiment, 505 individuals participated, with 8 respondents skipping a question and 10 providing incorrect answers. The gender distribution of incorrect responses followed a similar trend, with 72.2\% of the reactions coming from males and 27.8\% from females. These similar trends in both experiments suggest a gender-based difference in color vision, with men tending to have more color deficiencies than women. Our results also support the finding that males are more likely to be color blind, which aligns with the data reported by Colour Blind Awareness \cite{colourblindawarenessHome}. Overall, our findings suggest that approximately 1 in 19 individuals is color-blind (5.14\%).

\subsubsection{Outliers Analysis}

\begin{figure*}[ht!]
  \centering
  \makebox[\textwidth][c]{
 \includegraphics[width=0.9\textwidth]{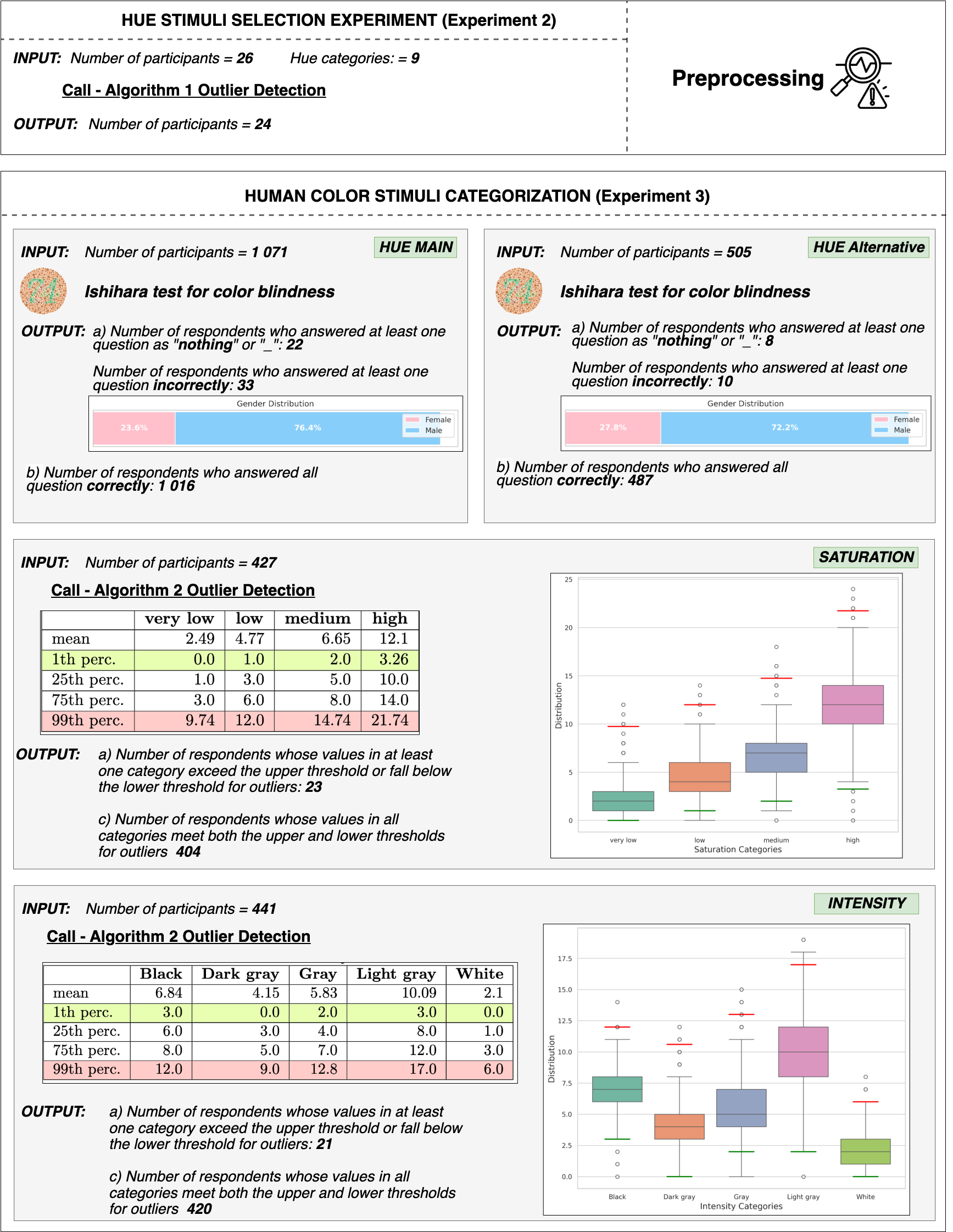}
  }
  \caption{Preprocessing}
  \label{fig:preprocessing}
\end{figure*}

In the preprocessing part of Experiment 2, illustrated in Fig. \ref{fig:preprocessing}, focused on \textit{Outlier Detection in Hue Stimuli Selection}, an Algorithm \ref{alg:hue_stumule_exp_2} was applied to analyze responses from 26 participants across nine color categories. This method involved calculating the interquartile range ($IQR$) to establish thresholds for outlier detection. Specifically, for each color category, responses were examined to identify values that exceeded the threshold defined by $Q3 + 1.5 \times IQR$, where these were initially marked as potential outliers. Participants whose proportion of outlier responses exceeded 40\% across all categories were subsequently classified as outliers.
Applying this algorithm resulted in the identification of two respondents as outliers. Consequently, their data were excluded from further analysis to ensure the integrity and accuracy of the study's findings.

\begin{algorithm}[ht!]
\caption{Outlier Detection in Hue Stimuli Selection
\\(Experiment 2)}
\label{alg:hue_stumule_exp_2}
\begin{algorithmic}[1]

\State \textbf{Input:} Responses from 26 participants across 9 color categories
\State \textbf{Output:} Identification of outliers

\Procedure{Calculate IQR Thresholds}{}
    \For{each color category}
        \State Calculate $Q1:quantile(0.25)$ 
        \State Calculate $Q3:quantile(0.75)$
        \State $IQR \gets Q3 - Q1$
        \State $Threshold \gets Q3 + 1.5 \times IQR$
        \For{each response in the category}
            \If{response $>$ Threshold}
                \State Mark response as a potential outlier
            \EndIf
        \EndFor
    \EndFor
\EndProcedure
\\
\Procedure{Outlier Conditions}{}
    \For{each participant}
        \State Initialize $count \gets 0$
        \For{each color category}
            \If {the response of the participant is marked as an outlier}
                \State $count \gets count + 1$
            \EndIf
        \EndFor
        \State $outlier\_percentage \gets (count / 9) \times 100$ 
        \If{$outlier\_percentage > 40$}
            \State Mark participant as an outlier
        \EndIf
    \EndFor
\EndProcedure

\State Call \textsc{Calculate IQR Thresholds}
\State Call \textsc{Outlier Conditions}
\end{algorithmic}
\end{algorithm}

In the preprocessing part of Experiment 3  - \textit{Outlier Detection in Saturation and Intensity Categorization}, we applied an Algorithm \ref{alg:satur_intens_exp_3} to identify outliers from the dataset, which included responses from participants on Saturation and Intensity Categorization. The algorithm calculated the 1st and 99th percentiles as thresholds to determine outliers for each attribute within the dataset. Responses that exceeded these established thresholds were marked as outliers. The results from the outlier detection algorithm showed that, in the saturation categorization task, 23 out of 427 respondents were identified as outliers. In the intensity categorization task, 21 out of 441 respondents were similarly identified. These outliers were excluded from further analysis to ensure the reliability and integrity of the study's findings. Detailed information about the preprocessing process and the results can be found in Figure \ref{fig:preprocessing}.

\begin{algorithm}
\caption{Outlier Detection in Saturation and Intensity Categorization (Experiment 3)}
\label{alg:satur_intens_exp_3}
\begin{algorithmic}[1]
\Function{calculate\_statistics}{df - Experiment responses}
    \For{column in df}
        \State $q01 \gets df[column].quantile(0.01)$
        \State $q99 \gets df[column].quantile(0.99)$
        \State $stats\_df[column] \gets [ q01, q99]$
    \EndFor
    \State \Return $stats\_df$
\EndFunction
\\
\Function{identify\_outliers}{df}
    \State $stats \gets calculate\_statistics(df)$

    \State $df['outlier'] \gets 0$
    \For{column in df}
        \State $upper\_threshold \gets stats['Q99'][column]$
        \State $lower\_threshold \gets stats['Q01'][column]$
        \State $df['outlier']  \gets (df[column] > upper\_threshold) \lor (df[column] < lower\_threshold)$
    \EndFor
    \State \Return $df$
\EndFunction
\end{algorithmic}
\end{algorithm}

\subsection{Fuzzy Partitions}

To guide the creation of fuzzy partitions, we visualized the survey data using scatter plots. This allowed us to observe and visually analyze how each class behaves and transitions across the range, which enabled us to determine whether a triangular or trapezoidal membership function should represent a class. Classes with distinct continuous cores were chosen to be trapezoidal, and classes with short or core peaking at one point were decided to be triangular.

After categorizing the membership types of each set, we search for all possible candidate partitions, which will be brute-forced and validated with experimental data to select the best-fitting partition with the least root mean square error across call partitions.


After obtaining our experimental survey data, we constructed fuzzy partitions for the color components, namely Hue, Saturation, and Intensity. The approach is universal for all color components. 

Firstly, for every color component, we use linguistic labels defined in the experiment.

For every linguistic class~\(C_j\) and for every sampling point~\(x_i\) we compute the empirical membership degree  
\[
\mu_{\text{emp}}(C_j,x_i)\;=\;\frac{\text{votes for }C_j\text{ at }x_i}{N}.
\]

Then, the distribution of the linguistic labels for the color component is analyzed using a scatter plot of sampling points and empirical membership from the experimental results.

Scatter plots of \((x,\mu_{\text{emp}})\) are inspected:
\begin{itemize}
  \item If the core covers a \emph{range}, we model \(C_j\) with a \emph{trapezoidal} set.
  \item If the core collapses to a \emph{peak}, we use a \emph{triangular} set.
\end{itemize}

By a qualitative decision, $C_j$ is assigned a fuzzy membership function form.

For parameterizing the partition, let the classes cover \(X\) in natural order with no gaps.  
A global cut‑point vector
\[
\mathbf{p}=(p_1,p_2,\dots,p_m),\qquad p_1<p_2<\dots<p_m,
\]
is sufficient to parameterize \emph{all} membership functions (triangular ones consume three consecutive points, trapezoidal ones four).  

\begin{table}[h!]
\centering
\caption{Generic mapping of classes to cut points and shapes.\label{tab:generic}}
\begin{tabular}{@{}lll@{}}
\toprule
Class & Parameters used & Shape \\
\midrule
$C_1$ & $\,\underline{L},\underline{L},p_1,p_2$ & trapezoid \\
$C_2$ & $p_1,p_2,p_3,p_4$ & trapezoid/triangle$^{\ast}$ \\
$\vdots$ & & \\
$C_k$ & $p_{m-1},p_m,\underline{U},\underline{U}$ & trapezoid \\
\bottomrule
\end{tabular}
\begin{flushleft}\small
$^{\ast}$Use three points if the class has a triangular shape, four otherwise.  
The underlined $L$ and $U$ are the lower and upper bounds of $X$.
\end{flushleft}
\end{table}

For a candidate \(\mathbf{p}\) we define
{\small
\begin{equation}
\mathrm{RMSE}(\mathbf{p}) =
\sqrt{\frac{1}{k} \sum_{j=1}^{k}
      \frac{1}{|X|} \sum_{x_i \in X}
      \left[\mu_{\text{model}}(C_j, x_i; \mathbf{p}) -
            \mu_{\text{emp}}(C_j, x_i)\right]^2}
\label{eq:rmse}
\end{equation}
}
The goal is to minimize \(\mathrm{RMSE}\).

To approximate fuzzy partitions based on experimental data, for each cut point \(p_\ell\), an empirically plausible interval \(I_\ell\) is specified. The possible candidates of $\mathbf{p}$ are formed with the Cartesian product \(I_1\times I_2\times\cdots\times I_m\) and evaluated with \(\mathrm{RMSE}(\mathbf{p})\) for every \(\mathbf{p}\) in that grid.

The vector with the smallest error is chosen. 



\begin{figure*}[htbp]
    \centering
    \begin{subfigure}{0.45\textwidth}
        \includegraphics[width=\linewidth]{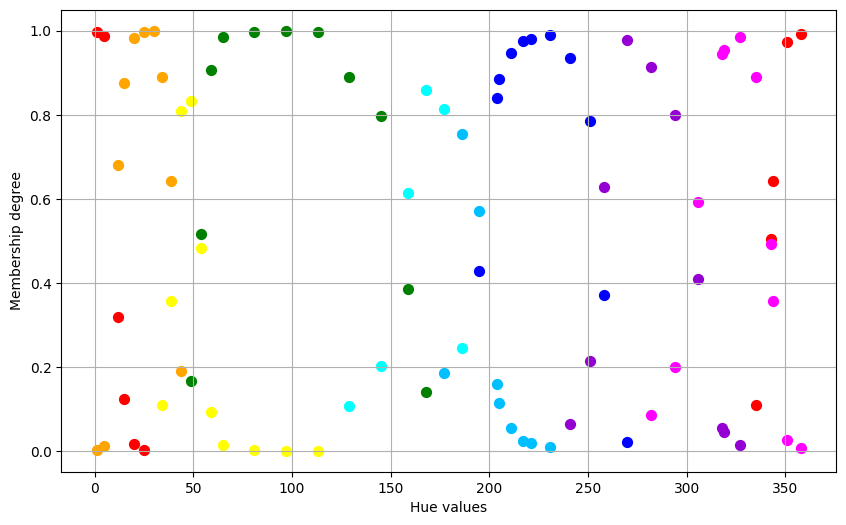}
        \caption{Hue scatter}
    \end{subfigure}
    \hfill
    \begin{subfigure}{0.45\textwidth}
        \includegraphics[width=\linewidth]{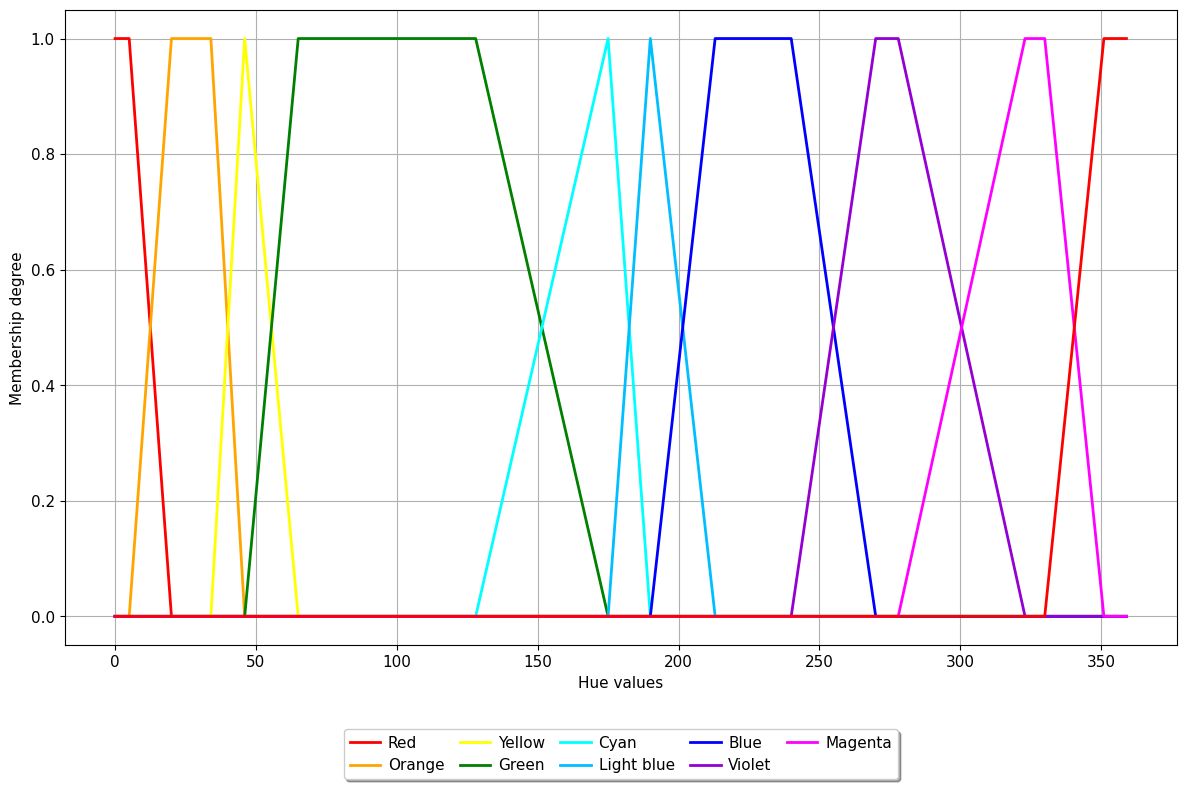}
        \caption{Hue fuzzy sets}
    \end{subfigure}

    
    
    \caption{Illustration of the fuzzy modelling workflow for the Hue color component: the (a) plots the raw hue, (b) fuzzy sets of hue, and (c) overlay of scatters on their respective fuzzy sets, revealing how the partitions capture the data distribution}
    \label{fig:3x1_grid}
\end{figure*}

\begin{figure*}
        \begin{subfigure}{0.5\textwidth}
    \includegraphics[width=\linewidth]{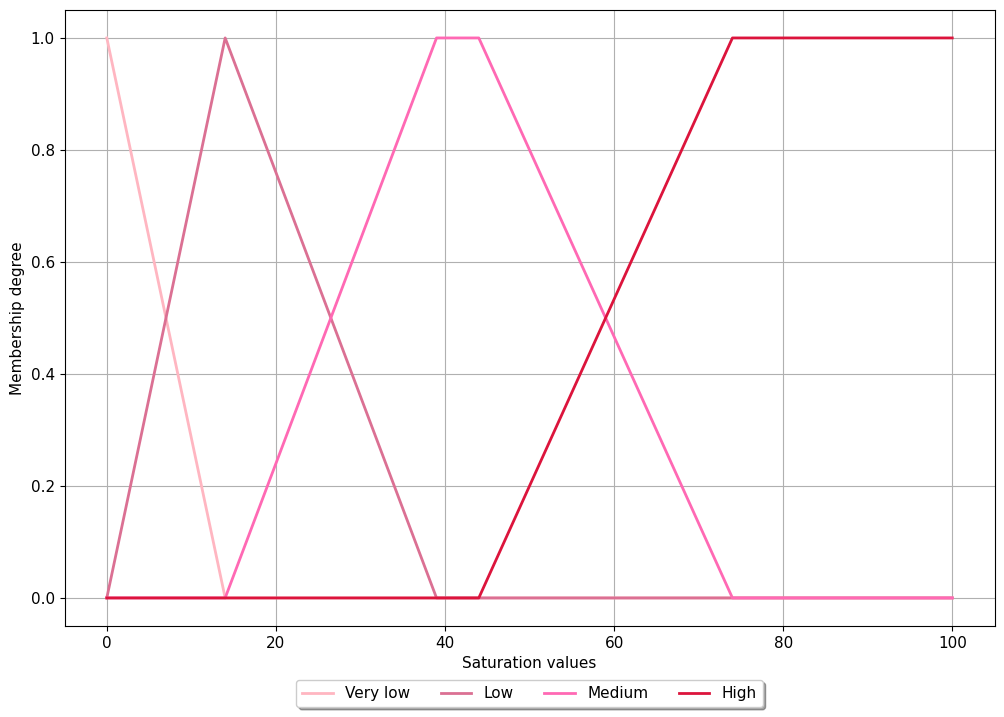}
    \caption{ Saturation fuzzy sets}
    \end{subfigure}
    \hfill
    \begin{subfigure}{0.5\textwidth}
        \includegraphics[width=\linewidth]{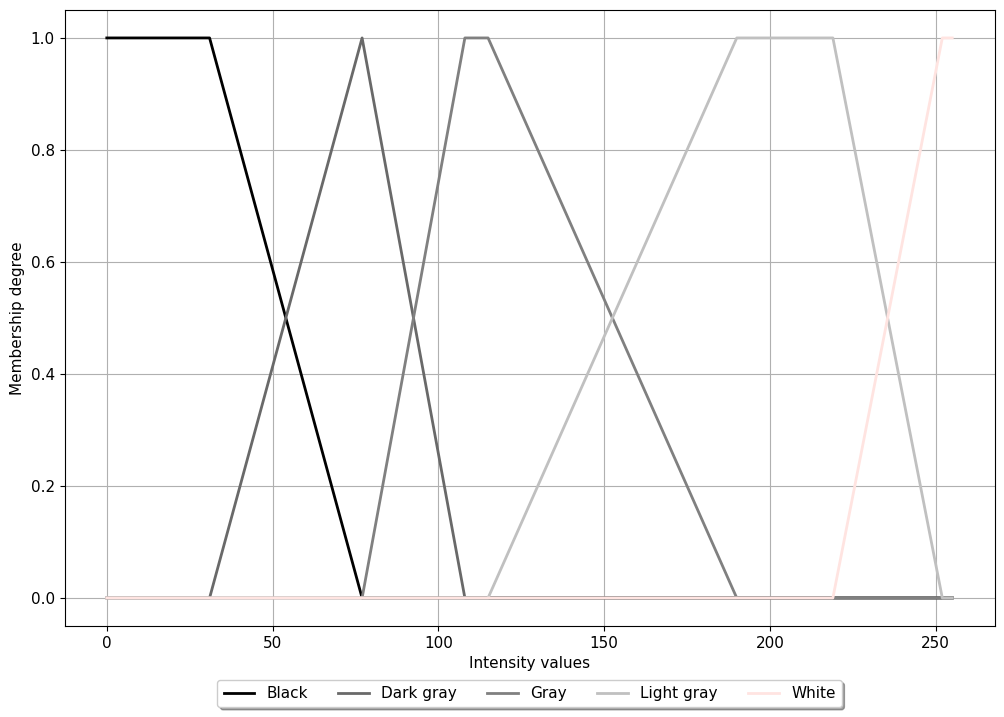}
        \caption{ Intensity fuzzy sets}
    \end{subfigure}
    \caption{Illustration of fuzzy sets of Saturation and Intensity.}
    \label{fuzzy_sets_int-sat}
\end{figure*}


\subsection{Performance Evaluation}
\subsubsection{Validation}



As mentioned in the previous sections, the hue categorisation analysis was done in two distinct ways. The results of the main survey serve as the foundation for our model membership functions. In contrast, the second survey is utilised to demonstrate the performance of the primary model as the validation metric. According to the results of the alternative hue categorisation experiment in Fig. \ref{alt-pie}, the results are mostly coherent with our main results. The coherence between the results of the main and hue experiments was calculated using Jensen-Shannon Divergence (JSD), Cosine Similarity, and Pearson Correlation. The quantitative results of this coherence assessment are presented in Table~\ref{tab:metric-interpretation}, where the numerical similarity between the two results of the survey is detailed. These results provide clear evidence of a high level of agreement between the survey formats, supporting the validity of the data collection process and the overall correctness of the survey conduct.

However, comparing the results of two hue categorisation surveys, we noticed a low coherence coefficient for stimuli 195 and 204, which were categorised into three hues – cyan, light blue, and blue – by respondents in two experiments. That is our primary challenge and a potential area for model enhancement.

Moreover, in this experiment, we provided respondents with multiple stimuli to choose from, allowing them to select from various options. In most cases, stimuli were categorized into two color categories, which supports our choice of providing two options in the main experiment for hue categorization.


\begin{table}[ht]
\centering
\begin{tabular}{|p{1.4cm}|p{0.7cm}|p{3cm}|p{1.4cm}|}
\hline
\textbf{Metric} & \textbf{Range} & \textbf{Interpretation} & \textbf{Coherence assessment} \\
\hline
\textbf{Jensen-Shannon Divergence (JSD)} & [0, 1] & 
0: Identical distributions \newline
1: Completely different & 0.1190 \\
\hline
\textbf{Cosine \newline Similarity} & [0, 1] & 
1: Maximum similarity \newline
0: Low similarity & 0.9348  \\
\hline
\textbf{Pearson \newline Correlation} & [-1, 1] & 
1: Perfect positive linear correlation \newline
0: No linear correlation \newline
-1: Perfect negative linear correlation  & 0.9245 \\
\hline
\end{tabular}
\caption{Correlation metrics.}
\label{tab:metric-interpretation}
\end{table}







\subsubsection{Example in Image Processing}

\begin{figure*}
  \centering

 \includegraphics[width=\textwidth]{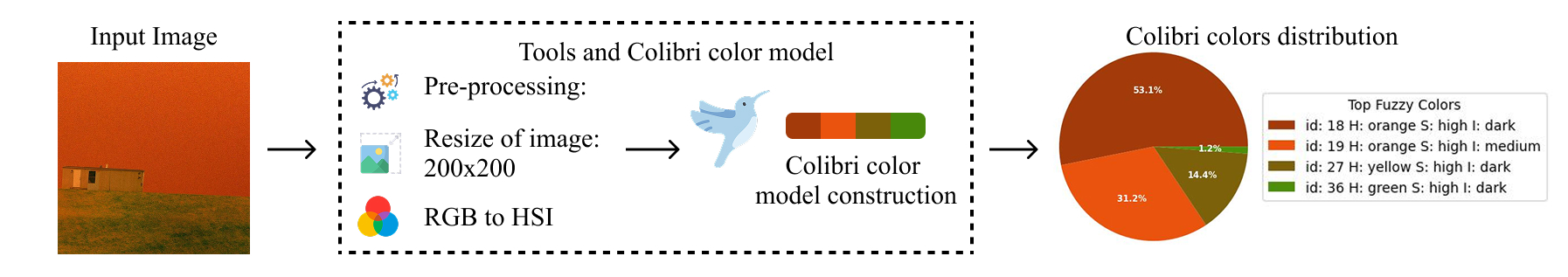}
  \caption{Dominant color extraction result} 
  \label{dominant_color_example}
\end{figure*}

\paragraph{Dominant color extraction}
Dominant color extraction is a common task in image processing that aims to identify the most prominent or perceptually important colors in an image \cite{Chang2022Color}. It typically returns a few primary colors with their proportions. These colors can then be used for recommendation or design. Common approaches include clustering algorithms, such as K-means. Some of the applications include color palette generation, image indexing or retrieval, computational aesthetics, and fashion \cite{Lai2020Machine}. Figure \ref{dominant_color_example} illustrates the model's performance in handling this task.

\paragraph{CBIR}

Content-Based Image Retrieval (CBIR) is a technique for retrieving images based on their visual content. It analyzes image features, including color, texture, shape, and spatial relationships, to measure the similarity between images. In contrast, Text-Based Image Retrieval (TBIR) relies on manually or automatically generated annotations, keywords, or captions to index and search images. These systems are widely used in various applications, including medical imaging, biometric identification, digital libraries, and e-commerce \cite{Swati2019, Iida2020}.

\begin{figure*}
  \centering
 \includegraphics[width=\textwidth]{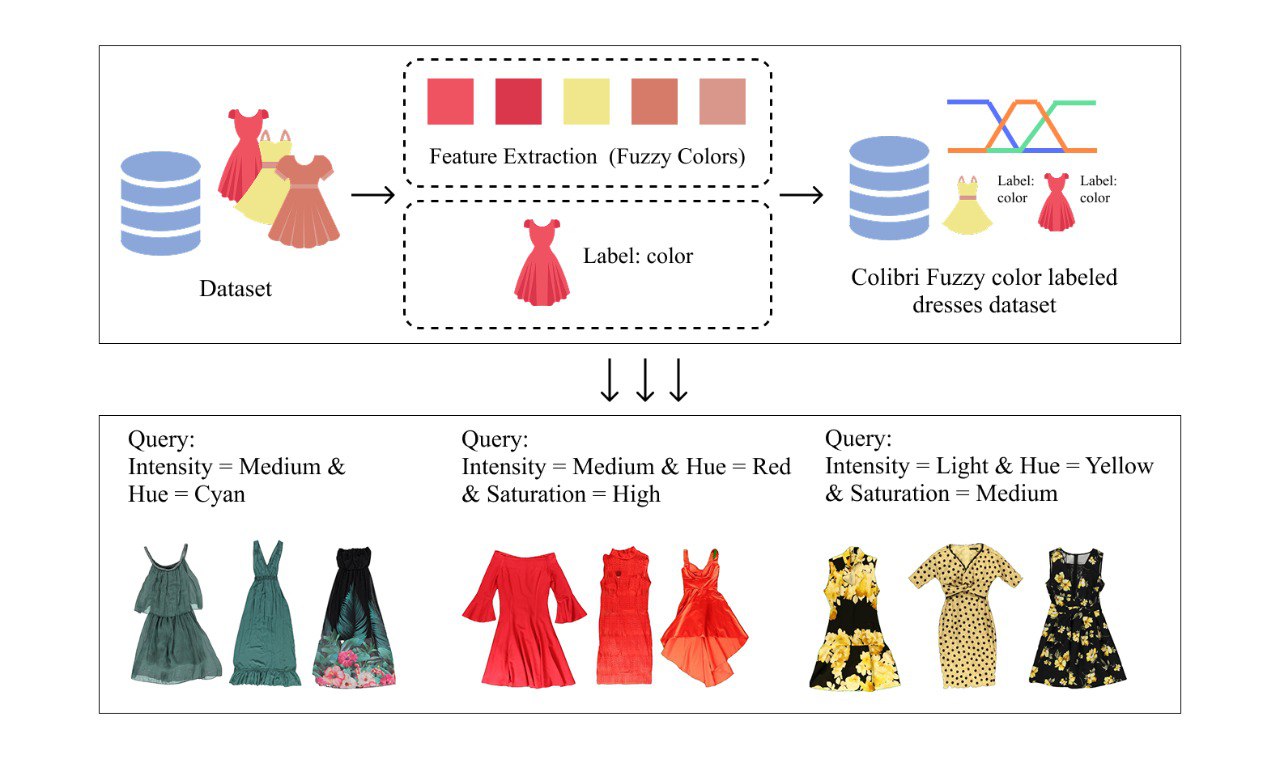}
  \caption{CBIR experiment result} 
  \label{cbir_example}
\end{figure*}

As an example of a possible application, we employed our model to label images in the dataset with a dominant color linguistic. For this experiment, we used a VISUELLE dataset \cite{Visuelle2021}. This resource contains 5577 fashion products described with multimodal information: image, textual metadata, sales, and popularity trends. For processing, we selected one category, specifically, dresses. The results can be observed in Figure \ref{cbir_example}.

\subsection{Agreement Evaluation}
To measure the consistency and reliability of the experimental results for the hue, saturation, and intensity surveys, we calculated Fleiss' Kappa agreement score for each survey mentioned. Fleiss' Kappa is defined in the following way:
\begin{equation}
\kappa = \frac{\bar{P} - \bar{P_e}}{1 - \bar{P_e}}
\label{eq:fleiss_kappa}
\end{equation}
 where: $\bar{P}$ = the mean of observed agreement over all items, $\bar{P_e}$ = the mean of expected agreement by chance across all categories.
 
 The following agreement scores were obtained: hue main - 0.76, saturation - 0.56, intensity - 0.49. Table 
\ref{tab:fleiss_kappa_interpretation} demonstrates the interpretation of Fleiss' Kappa values.

\begin{table}[h!]
\centering
\begin{tabular}{|c|l|}
\hline
\textbf{Kappa Value} & \textbf{Strength of Agreement} \\
\hline
$< 0.00$     & Poor \\
$0.00 - 0.20$ & Slight \\
$0.21 - 0.40$ & Fair \\
$0.41 - 0.60$ & Moderate \\
$0.61 - 0.80$ & Substantial \\
$0.81 - 1.00$ & Almost perfect \\
\hline
\end{tabular}
\caption{Interpretation of Fleiss' Kappa values}
\label{tab:fleiss_kappa_interpretation}
\end{table}






\section{Discussion}
\label{sec:discussion}
\subsection{Color Perception and Gender}
The study of \cite{Hurlbert2007} found that females showed a greater preference for warm colors, while males preferred cool colors, highlighting gender differences in color preferences.

Our findings align with prior research \cite{Bimler2004}, which has demonstrated that females exhibit greater sensitivity to hue variations, especially in the red-green and yellow-blue color axes. This supports the hypothesis that biological and cognitive factors influence color categorization at an individual level. The study also raises questions about linguistic influences on color categorization, as male and female participants may use different internal references when identifying hues. The results of \cite{Bimler2004} suggest that future studies on color naming, discrimination, and memory should consider sex-linked variations to develop more accurate models of human color processing.
\subsection{Comparison with recent studies}

These models are particularly important in fields such as artificial intelligence, design, and image processing. For example, the RGB model is used in computational color naming (CCN) to solve color vision deficiencies and to enhance image retrieval systems \cite{yan2022rgb}. Moreover, probabilistic color modeling is essential in fashion for extracting dominant colors from clothing items, aiding in recommender systems, and facilitating color trend analysis \cite{al2021probabilistic}. 

The study \cite{Elliot2014} examines the impact of color perception on human psychology. They discuss how different colors can carry specific meanings and elicit particular emotional and behavioral responses. Based on the framework established by \cite{Elliot2014}, this study examines the role of color in consumer decision-making processes.
\cite{Zhang2023} studied how artistic training affects color perception by analyzing both behavior and brain activity using event-related potentials (ERPs). Their research revealed that bright colors elicited stronger positive emotions and more intense brain responses, particularly in individuals with artistic training. This suggests that artistic training can improve the way people process visual information at a higher cognitive level.

Our findings align with \cite{Bornstein2007} model, which suggests that while perceptual mechanisms for color categorization are shared across populations, the linguistic boundaries and labels assigned to hues differ culturally. This study examines how color naming and categorization evolve from cognitive processes to language, influencing cultural interpretations of color.

Unlike traditional image enhancement techniques, such as the Multiscale Retinex algorithm \cite{Jobson1997}, which seeks to improve image clarity computationally, our study examines raw perceptual responses to color variations under natural viewing conditions.

\subsection{Adapting Fuzzy Sets:  Refining the Model with New Data}
To enhance the flexibility of the proposed Human Perception-based Color Model, an adaptive mechanism is necessary to refine categorization based on new datasets or survey feedback. As additional human-labeled data become available, the model can dynamically update fuzzy partitions and membership functions, thereby improving its alignment with real-world perception. This adaptation process can be achieved through machine learning techniques, where new survey responses are integrated to adjust existing color boundaries or introduce new categories. By continuously evolving with new datasets, the model can capture regional, cultural, or contextual variations in color perception, thereby enhancing its effectiveness in design, AI, and human-computer interaction.

\subsection{Perceptial boundary colors}


\begin{figure}[tb]
  \centering

 \includegraphics[width=0.3\textwidth]{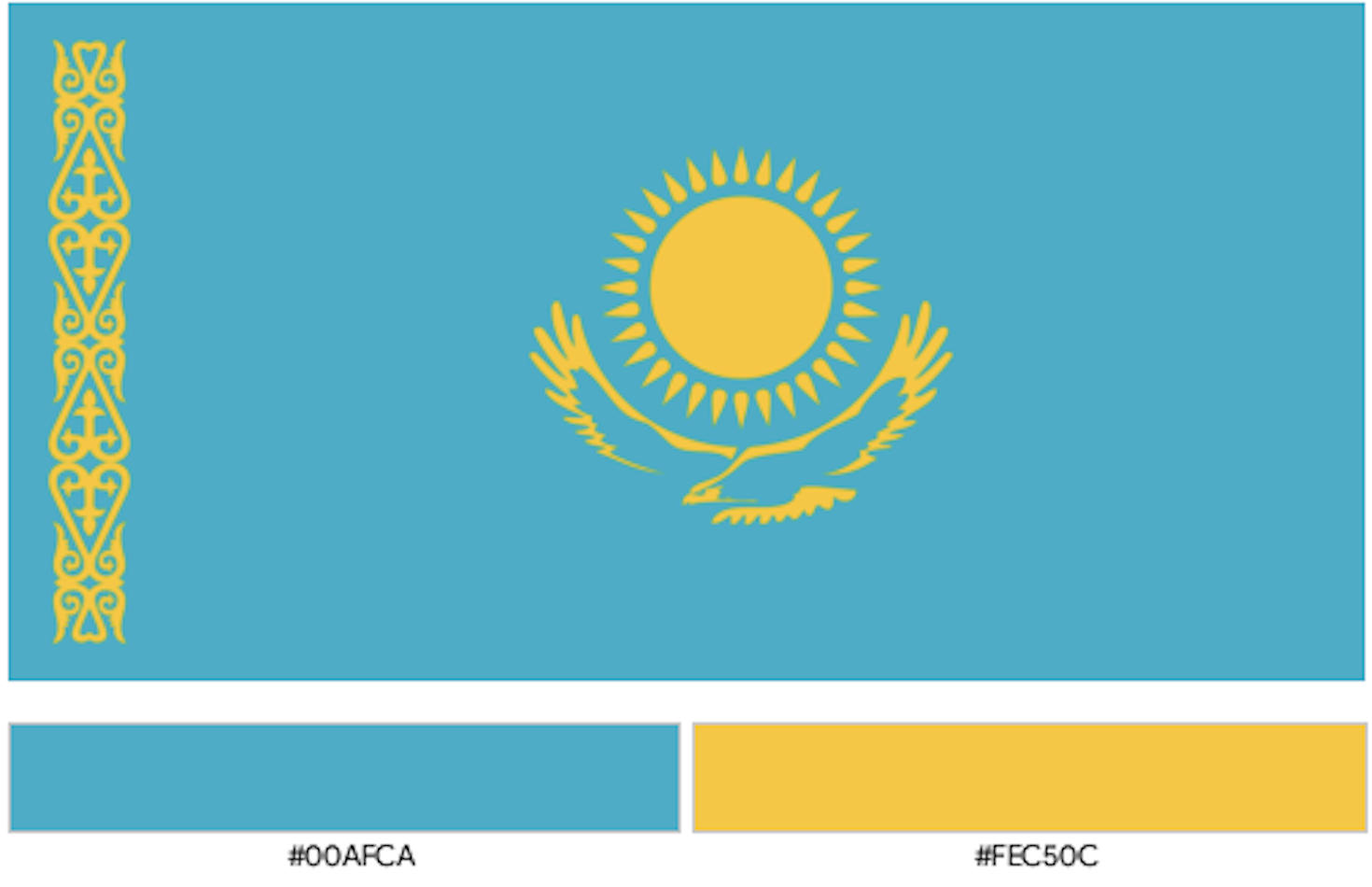}
  \caption[Caption for LOF]{The hex codes of colors in the Kazakhstan flag \protect\footnotemark}
  
  \label{flaghex}
\end{figure}

\footnotetext{https://www.flagcolorcodes.com/kazakhstan}

As discussed in the introduction, the color of the Kazakhstan flag can appear ambiguous—sometimes described as blue, cyan, turquoise, or even light blue, depending on the context, display, and language. While such inconsistencies are less frequent with other national flags, the Kazakhstan flag exemplifies what we define as a \textit{perceptual boundary color}. This color lies near the intersection of two fuzzy hue categories.

\begin{figure}
  \centering
 \includegraphics[width=0.5\textwidth]{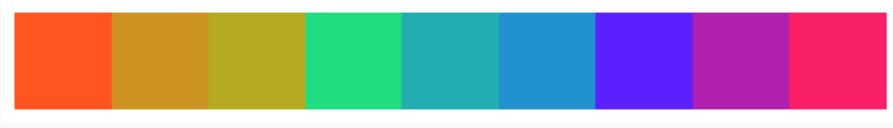}
  \caption{Boundary colors. From left to right: red-orange, orange-yellow, yellow-green, green-cyan, cyan-light blue, light blue-blue, blue-violet, violet-magenta, magenta-red} 
  \label{boundaries}
\end{figure}

\begin{figure}[tb]
 \includegraphics[width=0.5\textwidth]{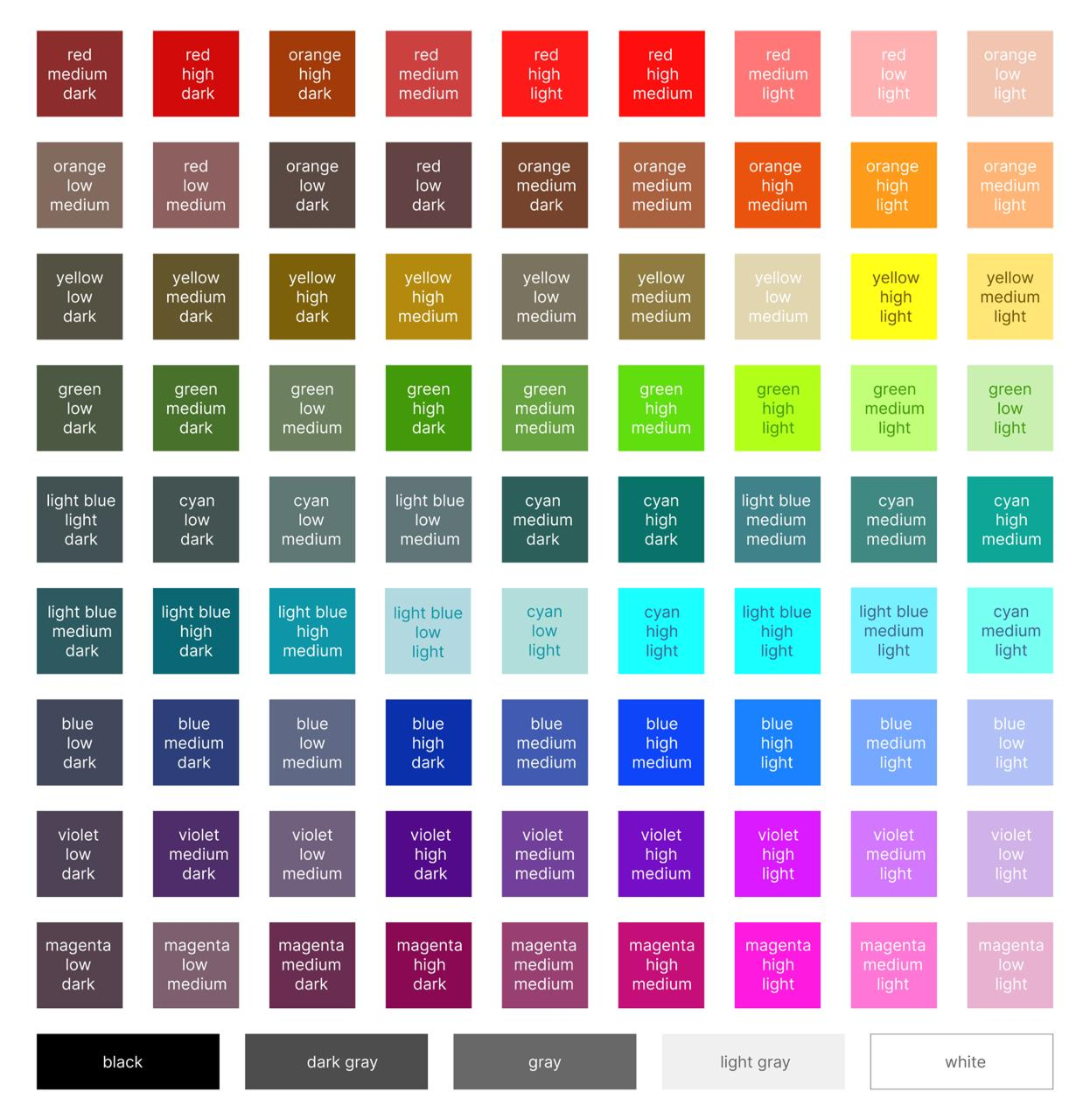}
 \caption{COLIBRI color model: obtained fuzzy colors. Each patch is fuzzy and represents a region, not a point, e.g., \textit{Hue = Red, Saturation = Medium, Intensity = Dark}}
 \label{color_palette}
\end{figure}

Through the construction of fuzzy partitions in our model, we identified specific hue regions where respondents were nearly equally divided in their classification—e.g., calling the same stimulus, either red or orange.  Fig. \ref{boundaries} illustrates several crisp representatives of such perceptual boundary colors extracted from our data.

Notably, the official background color of the Kazakhstan flag falls within the boundary region between \textit{cyan }and \textit{light blue} in our model (see Fig. \ref{flaghex}). Its measured HSI values (\textit{R=0, G=175, B=202} converted to HSI: \textit{H=187}) correspond to mixed membership degrees of 'Cyan' and  'Light blue' in the fuzzy sets we constructed. This result provides a perceptual explanation for the observed ambiguity: the flag's hue is inherently fuzzy and linguistically ambiguous, as it does not strictly belong to a single category. It supports the idea that the subjective, language-influenced nature of human color perception can explain why different viewers or reproduction systems interpret the same color differently.

This observation reinforces the practical value of fuzzy color models, particularly in explaining color categorization near perceptual boundaries, where linguistic variability and perceptual overlap make strict classification both difficult and unnatural.
\begin{table*}[ht!]
\centering
\renewcommand{\arraystretch}{1.5}
\caption{Summary of key findings.}
\begin{tabular}{p{0.8\textwidth} p{0.2\textwidth}}
\toprule
\textbf{Conclusion} & \textbf{Evidence} \\
\midrule
Certain hues—such as green, blue, and magenta—exhibit broader fuzzy sets. It indicates a wider range of human perception tolerance and recognition consistency. In contrast, colors like yellow, cyan, and light blue display narrower fuzzy sets. This variation in the wideness of the fuzzy sets reflects differences in how distinctly or ambiguously humans perceive each hue category & Fig.\ref{fig:3x1_grid} \\
Perceptual boundary colors & Fig.\ref{boundaries} \\
COLIBRI Fuzzy Model & Fig. \ref{fig:3x1_grid} \& Fig. \ref{fuzzy_sets_int-sat}\\
One of the first large-scale experimental efforts (n = 2496) to construct a perceptual color model grounded in human categorization of hue, saturation, and intensity. & Fig. \ref{fig:experiment} \\
Participants’ responses for each color stimulus in the main hue experiment: After removing outliers, the results show that most stimuli were clearly classified, with many (e.g., stimuli 1, 5, 20, etc.) receiving over 95\% agreement. A few showed minor misclassifications (85–95\%), while others (e.g., stimulus 251) had moderate disagreement (70–85\%). Some stimuli (e.g., 54, 343) received nearly equal responses, indicating ambiguous perception near category boundaries. & Fig. \ref{main-bar} \\
Participants’ responses in the saturation perception survey show consistent classification for most stimuli (1–5, 16–26), indicating clear perception. Stimuli 7 and 8 caused ambiguity between “low” and “medium” categories, possibly due to perceptual overlap or external factors. Overall, responses reflect strong agreement with a few areas of uncertainty. & Fig. \ref{fig:img-sat} \\
Participants’ responses in the intensity survey show overall agreement, especially at the beginning and end (black and white). Stimuli 7 and 11 caused ambiguity between adjacent intensity levels (e.g., black vs. dark gray), likely due to perceptual or display-related factors. & Fig. \ref{fig:img-int} \\
The Hue Main and Hue Alternative experiments of the Ishihara Test revealed that males had a higher rate of incorrect or skipped responses than females. These results indicate a gender-based difference in color vision, with approximately 1 in 19 people (5.14\%) being colorblind. &  Fig. \ref{fig:preprocessing} \\
The analysis showed minimal differences between male and female participants in color categorization, with mean values varying only slightly across color categories. Overall, the data suggest a shared conceptual understanding of color categories among participants of both genders within the study’s context.&  Tab. \ref{tab:hue_alter_stimule_count} \\
Previous research found that females are better at noticing color differences, especially between red-green and yellow-blue. Our results support this and suggest that biology and thinking affect how people categorize colors. &  Fig. \ref{main-bar} \& \ref{alt-pie}, Tab. \ref{tab:hue_alter_stimule_count} \\
High coherence between the main and alternative hue categorization surveys confirms the reliability of hue perception modeling, with minor deviations (e.g., stimuli 195 and 204) indicating potential for refinement. & Tab. \ref{tab:metric-interpretation} \\
The proposed model enables the effective annotation of dominant colors in fashion images, supporting its applicability in content-based image retrieval systems. & Fig. \ref{cbir_example} \\
\bottomrule
\end{tabular}
\end{table*}

\subsection{Open questions and challenges}
Following the experiment, several open questions and challenges remain, as color and human perception are complex and broad topics. One key question is whether color perception is consistent across people from different countries or if linguistic and cultural differences influence how colors are perceived. For instance, we encountered challenges distinguishing between cyan and light blue. Moreover, during the 'Hue Main' and 'Hue Alternative' experiments, we observed that participants responded differently to the same stimuli representing light blue, cyan, and blue colors.  Another area for further exploration is how the results and the color model might change under different experimental conditions or variations in the color stimuli.

To complement the quantitative data, we conducted semi-structured interviews with eight participants in the experiment to gain deeper insights into their linguistic and perceptual responses to color stimuli. The interview consisted of six open-ended questions, focusing on the ease or difficulty of describing specific stimuli, the perception of similar hues, experiment preferences, and the role of linguistic labels in color categorization:

\begin{enumerate}
    \item Which stimulus was the most difficult for you to describe linguistically?
    \item Which stimulus was the easiest for you to describe linguistically?
    \item Do you perceive a difference between the shades of blue, cyan, and light blue?
    \item Which hue experiment (main or alternative) was easier for you to complete, and why?
    \item Were there any questions you found difficult to answer or left unanswered?
    \item Did linguistic labels help you in choosing a color category?
\end{enumerate}

Notably, participants consistently identified colors such as yellow, magenta, and violet, as well as the distinction between cyan and light blue, as particularly challenging to describe linguistically. This supports the idea that certain hues exist near the boundaries of conventional color categories and may not correspond to widely accepted linguistic labels. This aligns with the fuzzy nature of color perception and supports the necessity of fuzzy membership functions in our COLIBRI model.

In contrast, primary colors such as red, green, and blue were reported as the easiest to recognize and label. The observed variation in ease of categorization across stimuli also emphasizes the importance of context and individual linguistic background, as some participants noted that familiarity with color names in their native language (e.g., Kazakh, Russian) improved their ability to label colors accurately.

In evaluating the hue experiments, participants showed divided preferences. Some favored the main experiment for its simplicity and focus on single-choice classification. In contrast, others noted the flexibility of the alternative experiment, which allowed them to express uncertainty by selecting multiple categories. This suggests that mixed experimental settings with different decision-making strategies may be beneficial in modeling real-world color perception.


\section{Conclusion}
In this paper, we introduced COLIBRI, an adaptive human perception-based color model that integrates fuzzy logic to better reflect the way humans name, perceive, and categorize colors. 

Unlike traditional color models such as RGB, HSV, and CIE-based models, which focus primarily on color matching across devices and rely on fixed numerical values, our approach incorporates linguistic categorization and perceptual variability, with soft transitions between color categories. COLIBRI is designed to be linguistically adaptive, making it better suited for real-world applications where colors are described and interpreted through language and cultural context rather than absolute numerical values.

The proposed model has a wide range of applications across multiple domains. In computer vision and AI, it can enhance color selection, object segmentation, and content-based image retrieval by assigning linguistic labels to colors in a manner that closely resembles human perception. In terms of design and aesthetics, COLIBRI can be used to predict color harmony, enhancing its application in branding, fashion, and architecture. Furthermore, in medical imaging, the model’s ability to categorize specific colors could help identify color variations in tissues, potentially aiding in the diagnosis of diseases.


The study has limitations.  Color perception is influenced by individual differences and cultural backgrounds, which may introduce variability in responses despite the fuzzy categorization approach. Additionally, while the model adapts to linguistic distinctions and human perception, it does not yet account for context-dependent color perception. So, as part of future work, we plan to incorporate more psychological aspects of perception (e.g., emotional impact of color) and adapt the model for specific fields (e.g., fashion, design, architecture). This study provides a scalable and reproducible framework for future research in human-centered color modeling and perceptually driven AI systems. 



\section*{Acknowledgement}
This research has been funded by the Science Committee of the Ministry of Science and Higher Education of the Republic of Kazakhstan (Grant No. AP22786412)

\bibliography{access}

\end{document}